\documentclass{article}

\usepackage{microtype}
\usepackage{graphicx}
\usepackage{subfigure}
\usepackage{booktabs} 

\usepackage{hyperref}

\usepackage[accepted]{icml2024}

\usepackage{amsmath}
\usepackage{amssymb}
\usepackage{mathtools}
\usepackage{amsthm}
\usepackage{multirow}
\usepackage{makecell}
\usepackage{float}
\usepackage{enumitem}

\usepackage[capitalize,noabbrev]{cleveref}

\theoremstyle{plain}

\theoremstyle{definition}

\theoremstyle{remark}

\usepackage[textsize=tiny]{todonotes}

\icmltitlerunning{UPOCR: Towards \underline{U}nified \underline{P}ixel-Level \underline{OCR} Interface}

\begin{document}

\twocolumn[
\icmltitle{UPOCR: Towards \underline{U}nified \underline{P}ixel-Level \underline{OCR} Interface}



\icmlsetsymbol{equal}{$\dagger$}

\begin{icmlauthorlist}
\icmlauthor{Dezhi Peng}{equal,scut,jointlab}
\icmlauthor{Zhenhua Yang}{equal,scut,jointlab}
\icmlauthor{Jiaxin Zhang}{scut,jointlab}
\icmlauthor{Chongyu Liu}{scut,jointlab}
\icmlauthor{Yongxin Shi}{scut,jointlab}
\icmlauthor{Kai Ding}{intsig,jointlab}
\icmlauthor{Fengjun Guo}{intsig,jointlab}
\icmlauthor{Lianwen Jin $^*$}{scut,jointlab}
\end{icmlauthorlist}

\icmlaffiliation{scut}{South China University of Technology}
\icmlaffiliation{intsig}{INTSIG Information Co. Ltd}
\icmlaffiliation{jointlab}{INTSIG-SCUT Joint Lab of Document Image Analysis and Recognition}

\icmlcorrespondingauthor{Lianwen Jin}{eelwjin@scut.edu.cn}

\icmlkeywords{Optical character recognition, Unified OCR Interface, Generalist Model, Text Removal, Text Segmentation, Tampered Text Detection}

\vskip 0.3in
]



\printAffiliationsAndNotice{\icmlEqualContribution} 

\begin{abstract}
Existing optical character recognition (OCR) methods rely on task-specific designs with divergent paradigms, architectures, and training strategies, which significantly increases the complexity of research and maintenance and hinders the fast deployment in applications.
To this end, we propose UPOCR, a simple-yet-effective generalist model for \underline{U}nified \underline{P}ixel-level \underline{OCR} interface.
Specifically, the UPOCR unifies the paradigm of diverse OCR tasks as image-to-image transformation and the architecture as a vision Transformer (ViT)-based encoder-decoder with learnable task prompts.
The prompts push the general feature representations extracted by the encoder towards task-specific spaces, endowing the decoder with task awareness.
Moreover, the model training is uniformly aimed at minimizing the discrepancy between the predicted and ground-truth images regardless of the inhomogeneity among tasks.
Experiments are conducted on three pixel-level OCR tasks including text removal, text segmentation, and tampered text detection.
Without bells and whistles, the experimental results showcase that the proposed method can simultaneously achieve state-of-the-art performance on three tasks with a unified single model, which provides valuable strategies and insights for future research on generalist OCR models.
Code is available at \url{https://github.com/shannanyinxiang/UPOCR}.
\end{abstract}

\section{Introduction}
\label{sec:intro}

Optical character recognition (OCR) is a flourishing field with numerous real-world applications, encompassing a wide spectrum of
pixel-level tasks that require dense per-pixel predictions, \textit{e.g.}, text removal~\cite{nakamura2017scene}, text segmentation~\cite{xu2021rethinking}, and tampered text detection~\cite{wang2022detecting}.
Nowadays, there have been massive methods that specialize in individual tasks, significantly contributing to the OCR advancement.

\begin{figure}[t]
    \centering 
    \includegraphics[width=0.9\columnwidth]{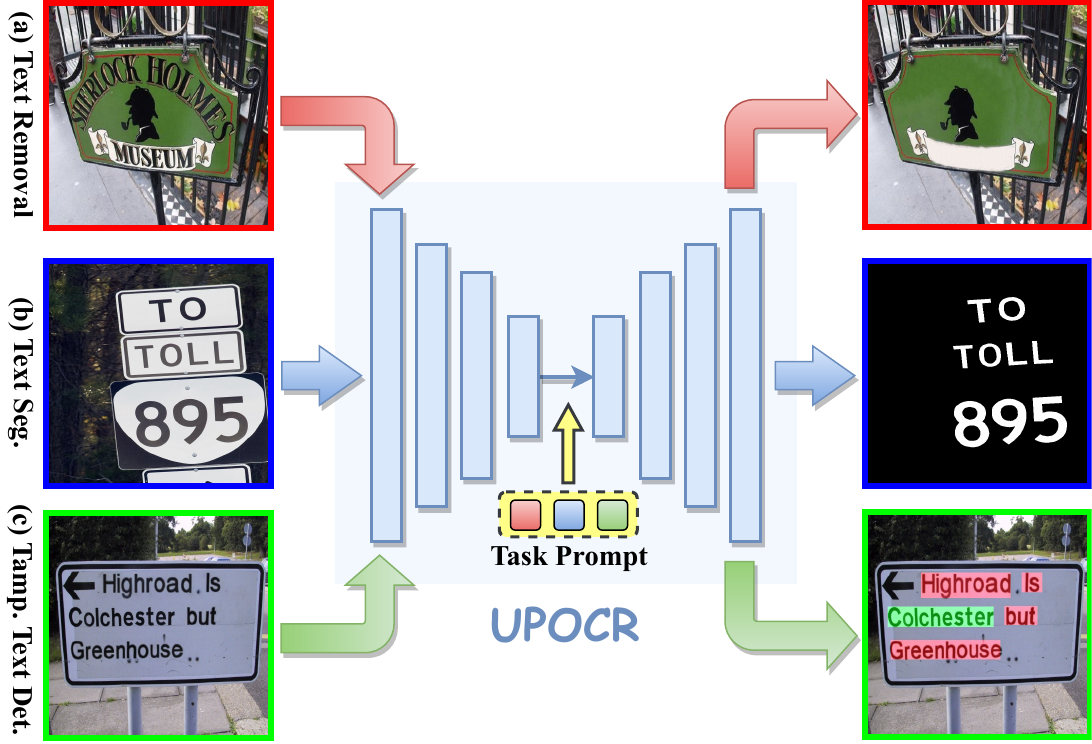}
    \caption{The proposed UPOCR is a unified pixel-level OCR interface which is simultaneously capable of diverse pixel-level OCR tasks (\textit{e.g.}, (a) text removal, (b) text segmentation, and (c) tampered text detection) by prompting ViT-based encoder-decoder, without task- or benchmark-specific finetuning.
    At the bottom right, red and green colors indicate tampered and real texts, respectively.}
    \label{fig:intro}
\end{figure}

However, specialized OCR models significantly differ in aspects of paradigms, architectures, and training strategies, especially for pixel-level OCR tasks.
\textbf{(1) Paradigm:}
The differences in paradigms are particularly reflected by divergent input and output formats.
For instance, existing approaches to text segmentation~\cite{xu2021rethinking} and tampered text detection~\cite{qu2023towards} typically transform the input image into per-pixel classification probabilities to distinguish text strokes or real/tampered texts from backgrounds.
On the contrary, text removal~\cite{zhang2019ensnet} is aimed at producing vivid text-erased images from input images.
Moreover, recent studies~\cite{liu2022don,lee2022surprisingly,wang2023pert} commonly consolidate text masks into inputs or outputs to promote text perception of text removal models.
\textbf{(2) Architecture:}
Specialized models typically elaborate dedicated modules for individual tasks.
Concretely, explicit text localization modules~\cite{tursun2020mtrnet++,du2023modeling} and multi-step refinements~\cite{liu2020erasenet,lyu2022psstrnet,wang2023pert} are employed to resolve excessive and inexhaustive text erasure.
As for text segmentation, advanced methods~\cite{bonechi2020weak,xu2021rethinking,ren2022looking,wang2023textformer} rely on complex attention and resampling modules as well as semantic information from text recognizers.
Moreover, the fusion of frequency and RGB domains is crucial to existing tampered text detection approaches~\cite{wang8tampered,qu2023towards}.
\textbf{(3) Training Strategy:}
Specialized models are optimized in disparate manners.
In general, the cross-entropy loss is adopted for text segmentation~\cite{xu2021rethinking,yu2023scene} and tampered text detection~\cite{qu2023towards} while L1 distance for text removal~\cite{zhang2019ensnet,liu2020erasenet}.
Additionally, abundant dedicated losses are designed for individual tasks, \textit{e.g.}, trimap loss~\cite{xu2021rethinking,xu2022bts} and Lovasz loss~\cite{qu2023towards}. 
Moreover, some studies~\cite{xu2021rethinking,xu2022bts,liu2022don,lyu2023fetnet,peng2024viteraser} incorporate discriminators and GAN-based training strategies.
The joint training with text localization modules is another trend in recent literature~\cite{wang2021semi,wang2023pert,yu2023scene}, requiring auxiliary supervision.
These inhomogeneities among specialized OCR models substantially raise the research complexity and increase the real-world deployment and maintenance cost.
More importantly, the collaboration between OCR tasks cannot be investigated.
Therefore, it is urgent to develop generalist pixel-level OCR models.

Concurrently, several studies have been devoted to establishing generalist interfaces.
However, OFA~\cite{wang2022ofa} and Unified-IO~\cite{lu2022unified} depend on VQGAN~\cite{esser2021taming} to decode images from discrete tokens, thus limiting the diversity and granularity in the pixel space.
Although recent approaches~\cite{alayrac2022flamingo,liu2023visual,li2023blip,ye2023mplugowl,zhang2024llama,zhu2024minigpt} investigate the combination of powerful vision Transformers (ViTs)~\cite{dosovitskiyimage} and large language models (LLMs)~\cite{touvron2023llama},
they struggle with OCR tasks~\cite{liu2023hidden,shi2023exploring} and fail to generate pixels.
Painter~\cite{wang2023images} tackles diverse tasks as inpainting problems but implicitly distinguishes tasks via example pairs, hence difficult to identify OCR tasks with inconspicuous correlation between inputs and outputs.
Furthermore, existing generalist OCR models~\cite{kim2022ocr,tang2023unifying,blecher2023nougat,lv2023kosmos,ye2023ureader,feng2023unidoc,ye2023mplug,feng2023docpedia} primarily focus on document scenarios and cannot handle pixel-level tasks.
In particular, some of them~\cite{kim2022ocr,tang2023unifying} still require benchmark-specific finetuning to reach satisfactory performance.

To this end, we propose UPOCR, a simple-yet-effective generalist model for \underline{U}nified \underline{P}ixel-level \underline{OCR} interface.
For the first time, UPOCR simultaneously accomplishes multiple pixel-level OCR tasks via prompting ViT-based encoder-decoder as shown in Fig.~\ref{fig:intro}.
To achieve this, the proposed method unifies the paradigm, architecture, and training strategy of diverse pixel-level OCR tasks.
Specifically, UPOCR unified the paradigm of different tasks as transforming RGB image inputs to RGB image outputs.
Moreover, the pure ViT-based encoder-decoder architecture is uniformly adopted for all tasks.
To distinguish the ongoing task, we introduce learnable task prompts into the encoder-decoder.
Concretely, the encoder first extracts general OCR-related feature representations of the input image. 
Subsequently, the task prompt pushes the general feature towards the task-specific region, empowering the decoder to generate output images for specific tasks.
During training, the model is optimized to minimize the discrepancy between predicted and ground-truth (GT) images at pixel and feature levels, eliminating dedicated designs of loss functions, adversarial learning, and auxiliary supervision.

The effectiveness of UPOCR is extensively verified on three pixel-level OCR tasks, including text removal, text segmentation, and tampered text detection.
Without bells and whistles, experimental results showcase that UPOCR simultaneously achieves state-of-the-art performance on all three tasks with a unified single model, significantly surpassing specialized methods for individual tasks.
In addition, UPOCR outperforms cutting-edge generalist approaches on these three tasks, demonstrating its proficiency in the pixel-level OCR field.

In summary, the contributions of this paper are as follows. 
\begin{itemize}[itemsep=1pt,topsep=0pt,parsep=0pt]
    \item We propose UPOCR, a simple-yet-effective generalist model for unified pixel-level OCR interface.
    Through the unification of paradigms, architectures, and training strategies, the proposed UPOCR is the first to simultaneously excel in diverse pixel-level OCR tasks.
    \item Learnable task prompts are introduced to guide the ViT-based encoder-decoder architecture.
    The prompts push general feature representations from the encoder towards regions of individual tasks, allowing the decoder to perform task-specific decoding. 
    \item The generalist capacity of UPOCR is extensively verified on text removal, text segmentation, and tampered text detection tasks, significantly outperforming existing specialized models. 
    In-depth ablation studies are also conducted to provide valuable strategies and insights for future research on generalist OCR methods. 
\end{itemize}

\section{Related Work}
\subsection{Specialized Pixel-Level OCR Model}
\noindent\textbf{Text Removal.}
Text removal is targeted at replacing text strokes with visually coherent backgrounds, primarily focusing on natural scenes.
Early approaches~\cite{nakamura2017scene,zhang2019ensnet,liu2020erasenet} follow a one-stage framework, which implicitly integrates text localization and inpainting processes into a single network in an image-to-image translation manner.
However, one-stage approaches struggle with accurate text perception, leaving severe text remnants in text removal results.
Therefore, two-stage methods incorporate explicit text segmentation modules~\cite{tursun2020mtrnet++,keserwani2021text,lyu2022psstrnet,bian2022scene,du2022progressive,hou2022multi,du2023modeling,wang2023pert,lyu2023fetnet} or external text detectors~\cite{zdenek2020erasing,conrad2021two,liu2022don,qinautomatic,tursun2019mtrnet,tang2021stroke,lee2022surprisingly} for enhanced text localization capacity and have recently dominated the text removal field.
Moreover, coarse-to-refine~\cite{liu2020erasenet,jiang2022self,tursun2020mtrnet++,du2023modeling} and multi-step progressive refinements~\cite{lyu2022psstrnet,bian2022scene,du2022progressive,wang2023pert} have also been intensively exploited for exhaustive text removal in recent studies.
Nevertheless, ViTEraser~\cite{peng2024viteraser} demonstrated a one-stage framework with ViTs and SegMIM pre-training can significantly outperform previous complicated methods.

\noindent\textbf{Text Segmentation.}
Text segmentation aims to predict per-pixel classification for distinguishing text strokes from backgrounds.
SMANet~\cite{bonechi2020weak} inserted multi-scale attention into PSPNet~\cite{zhao2017pyramid}-based architecture.
TextFormer~\cite{wang2023textformer} and ARM-Net~\cite{ren2022looking} further incorporated high-level semantics from text recognizers.
Moreover, TexRNet~\cite{xu2021rethinking} dynamically reactivated low-confidence regions and adopted a character-level discriminator.
Based on TexRNet, PGTSNet~\cite{xu2022bts} additionally integrated a text detector for text-line cropping and employed a line-level discriminator.
To fully utilize polygon annotations, \citet{wang2021semi} exploited mutual interaction between polygon- and pixel-level segmentation while \citet{yu2023scene} designed end-to-end hierarchical segmentation Transformers.

\noindent\textbf{Tampered Text Detection.}
Tampered text detection is defined as the segmentation (or detection) of tampered (or both real and tampered) texts.
In the deep learning era, plenty of approaches~\cite{zhou2018learning,bappy2019hybrid,kwon2021cat,kwon2022learning,dong2022mvss} have been developed for natural image manipulation detection.
Inspired by these methods, \citet{qu2023towards} proposed DTD which combines frequency and visual perception for document tampered text detection.
Furthermore, \citet{wang8tampered} enhanced Faster R-CNN with RGB and frequency relationship modeling for tampered text detection in receipts.
As for natural scenes, \citet{wang2022detecting} equipped scene text detectors with the proposed S3R strategy to localize both real and tampered texts.

\subsection{Generalist Model}
The emergence of Transformer~\cite{vaswani2017attention,dosovitskiyimage} breaks the boundary between different modalities~\cite{jaegle2022perceiver,zhang2023meta}, fostering a broad variety of generalist models.
One category of these models unifies both the input and output as sequences and bridges them with a \textit{sequence-to-sequence} learning framework.
Pix2Seq~\cite{chen2022pix2seq} pioneered in unifying the output vocabulary of natural language and spatial coordinates and demonstrated effectiveness in object detection.
Subsequently, Pix2Seq v2~\cite{chen2022unified} simultaneously tackled multiple tasks with the guidance of specific prompts.
Furthermore, OFA~\cite{wang2022ofa} discretized both the input and output as token sequences, accomplishing various uni-modal and cross-modal vision-language tasks. 
Similarly, Unified-IO~\cite{lu2022unified} extended the framework to a wider range of tasks and modalities. 
With the rise of LLMs~\cite{openai2023gpt4,touvron2023llama}, numerous studies~\cite{alayrac2022flamingo,li2023blip,liu2023visual,ye2023mplugowl,zhang2024llama,zhu2024minigpt} connected pretrained ViTs and LLMs for generalist models with stronger reasoning and robustness.
Following an \textit{image-to-image} translation pipeline, MAE-VQGAN~\cite{bar2022visual} treated diverse tasks as inpainting problems.  
Painter~\cite{wang2023images} further investigated visual in-context learning which allows it to adapt to unseen tasks. 


In the OCR field, several generalist models have been studied following \textit{sequence-to-sequence} paradigms~\cite{kim2022ocr,tang2023unifying,blecher2023nougat,lv2023kosmos}.
Moreover, recent approaches~\cite{ye2023ureader,ye2023mplug,feng2023unidoc,feng2023docpedia} augment large multimodal models using OCR-related data and fine-grained visual perception.
However, these approaches primarily focused on document scenarios and failed to generate pixels.
Moreover, some of them~\cite{tang2023unifying,kim2022ocr} still required benchmark-specific finetuning.

\begin{figure*}[t]
    \centering 
    \includegraphics[width=2\columnwidth]{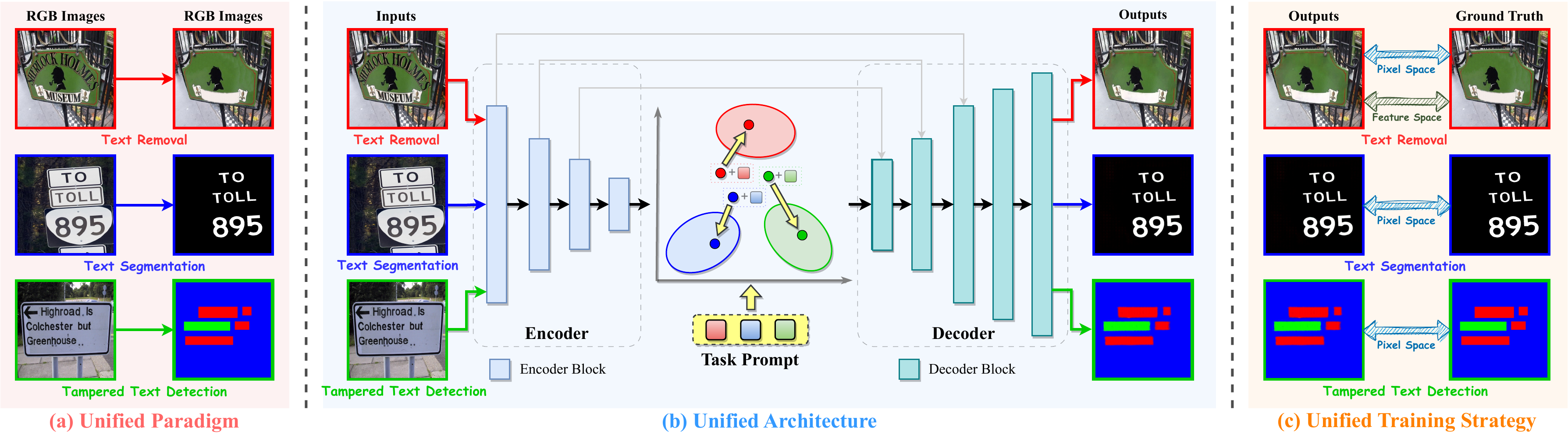}
    \caption{UPOCR is a generalist OCR model which unifies the paradigm, architecture, and training strategy of diverse pixel-level OCR tasks. 
    (a) The paradigm is unified as RGB image to RGB image translation.
    (b) The ViT-based encoder-decoder architecture is employed for all tasks. 
        Learnable task prompts are inserted to shift general hidden representations to task-specific regions.
    (c) During training, the model is optimized to minimize the discrepancy between predicted and GT images at pixel and feature spaces.}
    \label{fig:method}
\end{figure*}

\section{Methodology}

The proposed UPOCR is a unified pixel-level OCR interface as depicted in Fig.~\ref{fig:method}.
Specifically, UPOCR unifies the paradigm, architecture, and training strategy of diverse pixel-level OCR tasks.
In this paper, the UPOCR is particularly verified on simultaneously handling text removal, text segmentation, and tampered text detection tasks. 

\subsection{Unified Paradigm}
\label{sec:method_paradigm}
As illustrated in Fig.~\ref{fig:method}(a), despite their divergent targets (\textit{e.g.}, image generation and segmentation), the paradigm of various pixel-level OCR tasks can be unified to translate RGB images to RGB images. 
As the inputs are inherently RGB images, detailed output formats of selected tasks are described as follows.

\noindent\textbf{Text Removal.}
For the text removal task, the output is the text-erased image corresponding to the input, which is also RGB pictures.

\noindent\textbf{Text Segmentation.}
Text segmentation aims to assign each pixel to foreground (\textit{i.e.}, text stroke) or background.
Existing methods typically conduct per-pixel binary classification.
However, under the unified image-to-image translation paradigm, UPOCR predicts RGB images with white and black colors.
Specifically, the RGB values for foreground and background pixels are $(255, 255, 255)$ and $(0, 0, 0)$, respectively.
During inference, the category is determined by thresholding the distance between the generated RGB value and the pre-defined foreground RGB value.

\noindent\textbf{Tampered Text Detection.}
Following recent studies~\cite{wang2022detecting}, we define tampered text detection as per-pixel classification of tampered text, real text, and background categories.
Similar to text segmentation, we adopt different RGB values in the output image to represent different categories.
Concretely, we assign red $(255, 0, 0)$, green $(0, 255, 0)$, and blue $(0, 0, 255)$ colors to tampered texts, real texts, and backgrounds, respectively.
During inference, we compare the distance of predicted RGB values with these three colors to determine the per-pixel category.

\subsection{Unified Architecture}
\label{sec:method_UA}
As shown in Fig.~\ref{fig:method}(b), we implement the unified image-to-image translation paradigm by prompting a ViT-based encoder-decoder network.
Concretely, the task prompts shift the general representation extracted by the encoder into task-specific regions at feature space, empowering the decoder to produce output images for individual tasks.

\noindent\textbf{Encoder-Decoder.}
We adopt a ViT-based architecture to implement the encoder-decoder of UPOCR.
Specifically, the encoder consists of four sequential blocks, yielding four feature maps with strides of $\{4, 8, 16, 32\}$ \textit{w.r.t} the input image.
Each encoder block encapsulates a patch embedding layer for downsampling and multiple Swin Transformer v2 blocks~\cite{liu2022swin}.
Subsequently, the decoder hierarchically upsamples the final feature of the encoder to strides of $\{16, 8, 4, 2, 1\}$ \textit{w.r.t} the input size through five blocks. 
Each decoder block concatenates multiple Swin Transformer v2 blocks~\cite{liu2022swin} and a patch splitting layer~\cite{peng2024viteraser} for upsampling.
Based on the final feature of the decoder, the output image is predicted using a $3 \times 3$ convolution.

\noindent\textbf{Task Prompt.}
To effectively handle multiple tasks, we introduce learnable task prompts into the encoder-decoder architecture.
Retrospectively, recent generalist models~\cite{chen2022unified,kim2022ocr,tang2023unifying} commonly prepend task prompts comprising natural language or pre-defined tokens to the decoder for task-specific sequence generation.
In contrast, we insert learnable task prompts into the shared feature space of the encoder and decoder as shown in Fig.~\ref{fig:method}(b).
Specifically, for each task, its prompt is formulated as a learnable embedding with the same dimension as the hidden feature from the encoder.
To perform a certain task, the corresponding prompt is simply added to every pixel of the hidden feature, pushing the general OCR-related presentations generated by the encoder towards the task-specific region.
Subsequently, the decoder translates the adjusted hidden feature into the output image for this specific task.
With negligible parameter and computation overhead, the UPOCR can simultaneously deal with diverse tasks in a simple yet effective fashion.
See Sec.~\ref{sec:appendix_network_architecture} of the appendix for details.

\begin{table*}[t]
    \centering 
    \caption{Ablation study on the insertion of task prompts. 
    The \textbf{bold} and \underline{underline} indicate the best and second best, respectively.
    }
    \label{tab:prompt_insertion}
    \resizebox{1.6\columnwidth}{!}{
        \begin{tabular}{ccccccccccc}
            \toprule
            \multicolumn{3}{c}{Prompt Insert Position} & \multicolumn{4}{c}{Text Removal} & \multicolumn{2}{c}{Text Segmentation} & \multicolumn{2}{c}{Tampered Text Det.} \\
            \cmidrule(r){1-3} \cmidrule(r){4-7} \cmidrule(r){8-9} \cmidrule(r){10-11}
            Encoder & Shared Feature & Decoder & PSNR$\uparrow$ & MSSIM$\uparrow$ & MSE$\downarrow$ & FID$\downarrow$ & fgIoU$\uparrow$ & F$\uparrow$ & mIoU$\uparrow$ & mF$\uparrow$ \\
            \midrule
            $\checkmark$ & $\checkmark$ & & 36.93 & \textbf{97.64} & \underline{0.0430} & \textbf{10.45} & 88.61 & 93.96 & 70.19 & 82.48 \\
            & $\checkmark$ & $\checkmark$ & \underline{36.96} & \textbf{97.64} & \textbf{0.0428} & \underline{10.46} & \underline{88.78} & \underline{94.06} & 71.13 & 83.13 \\
            $\checkmark$ & $\checkmark$ & $\checkmark$ & 36.86 & \textbf{97.64} & 0.0436 & 10.55 & \textbf{88.84} & \textbf{94.09} & \underline{71.40} & \underline{83.31} \\
            \midrule
            & $\checkmark$ & & \textbf{37.14} & \underline{97.62} & \textbf{0.0428} & 10.47 & 88.76 & 94.04 & \textbf{71.71} & \textbf{83.53} \\
            \bottomrule
        \end{tabular}
    }
\end{table*}


\subsection{Unified Training Strategy}
\label{sec:unified_training_strategy}
Thanks to the unified image-to-image paradigm, the training of UPOCR consistently aims to minimize the discrepancy between the predicted and GT images at pixel and feature spaces regardless of the inhomogeneity among tasks, as shown in Fig.~\ref{fig:method}(c).
For conciseness, we define the input, output, and GT images as $I_{in}$, $I_{out}$, and $I_{gt}$, respectively.

\noindent\textbf{Pixel Space.}
The discrepancy in pixel space is measured by the L1 distance between output and GT images. 
Furthermore, a multi-scale L1 loss is involved to enhance the perception at multiple granularities during training.
Specifically, multi-scale images $I_{out}^{\frac{1}{4}}$ and $I_{out}^{\frac{1}{2}}$ are predicted based on the features from the 3\textit{rd} and 4\textit{th} decode blocks, each through a $3\times3$ convolution.
Denoting $\mathbb{I}_{out} = \{I_{out}^{\frac{1}{4}}, I_{out}^{\frac{1}{2}}, I_{out}\}$ and $\mathbb{I}_{gt} = \{I_{gt}^{\frac{1}{4}}, I_{gt}^{\frac{1}{2}}, I_{gt}\}$ ($I_{gt}^{\frac{1}{4}}$ and $I_{gt}^{\frac{1}{2}}$ is resized from $I_{gt}$), the pixel loss is formulated as:
\begin{equation}
    L_{pix} = \sum\nolimits_{i=1}^3 \alpha_i || \mathbb{I}_{out}^{i} - \mathbb{I}_{gt}^{i} ||_1,
    \label{equ:pixel_loss}
\end{equation}
where the balancing factor $\alpha$ is empirically set to $\{5, 6, 10\}$ and the superscript $i$ indicates the $i$-th element of the array.

\noindent\textbf{Feature Space.}
For tasks associated with realistic image generation, the similarity at high-level feature space is critical.
Therefore, we additionally align the output and GT images at the feature space for the text removal task.
The feature loss $L_{feat}$ is composed of perceptual loss $L_{per}$~\cite{johnson2016perceptual} and style loss $L_{sty}$~\cite{gatys2016image}:
\begin{equation}
    L_{feat} = \alpha_{per} L_{per} + \alpha_{sty} L_{sty},
    \label{equ:feature_loss}
\end{equation}
where $L_{per}$ and $L_{sty}$ are calculated following previous studies~\cite{liu2020erasenet,liu2022don} using a pretrained VGG-16~\cite{Simonyan15} network. In addition, the $\alpha_{per}$ and $\alpha_{sty}$ are heuristically set to 0.01 and 120, respectively.

\noindent\textbf{Total Loss.}
The total loss $L_{total}$ is the sum of pixel loss $L_{pix}$ and feature loss $L_{feat}$ (if applicable):
\begin{equation}
    L_{total} = L_{pix} + L_{feat}.
\end{equation}

    
\begin{table*}[t]
    \centering 
    \caption{Ablation study on task collaboration. 
    (TR: text removal, TS: text segmentation, TTD: tampered text detection)
    }
    \label{tab:task_collaboration}
    \resizebox{1.28\columnwidth}{!}{
    \begin{tabular}{ccccccccccc}
        \toprule
        \multirow{2}*{TR} & \multirow{2}*{TS} & \multirow{2}*{TTD} & \multicolumn{4}{c}{Text Removal} & \multicolumn{2}{c}{Text Segmentation} & \multicolumn{2}{c}{Tampered Text Det.} \\
        \cmidrule(r){4-7} \cmidrule(r){8-9} \cmidrule(r){10-11}
        & & & PSNR$\uparrow$ & MSSIM$\uparrow$ & MSE$\downarrow$ & FID$\downarrow$ & fgIoU$\uparrow$ & F$\uparrow$ & mIoU$\uparrow$ & mF$\uparrow$ \\
        \midrule
        $\checkmark$ & & & 36.97 & 97.57 & 0.0500 & 10.50 & - & - & - & - \\
        & $\checkmark$ & & - & - & - & - & \underline{88.83} & \underline{94.08} & - & - \\
        & & $\checkmark$ & - & - & - & - & - & - & 67.73 & 80.72 \\
        \midrule
        $\checkmark$ & $\checkmark$ & & 36.91 & \textbf{97.64} & \underline{0.0450} & \textbf{10.40} & 88.64 & 93.98 & - & - \\
        & $\checkmark$ & $\checkmark$ & - & - & - & - & \textbf{89.07} & \textbf{94.22} & \underline{70.78} & \underline{82.89} \\
        $\checkmark$ & & $\checkmark$ & \underline{37.08} & 97.60 & 0.0452 & \underline{10.47} & - & - & 69.26 & 81.83 \\
        \midrule
        $\checkmark$ & $\checkmark$ & $\checkmark$ & \textbf{37.14} & \underline{97.62} & \textbf{0.0428} & \underline{10.47} & 88.76 & 94.04 & \textbf{71.71} & \textbf{83.53} \\
        \bottomrule
    \end{tabular}
    }
\end{table*}

\section{Experiment}
\subsection{Experiment Setting}
\label{sec:exp_set}

\noindent\textbf{Task and Dataset.}
We investigate three pixel-level OCR tasks, including text removal, text segmentation, and tampered text detection, to demonstrate the effectiveness of UPOCR on generalist pixel-level OCR processing.
The SCUT-EnsText~\cite{liu2020erasenet}, TextSeg~\cite{xu2021rethinking}, and Tampered-IC13~\cite{wang2022detecting} datasets are employed for these three tasks, respectively.

\noindent\textbf{Network Architecture.}
The encoder-decoder architecture of UPOCR inherits from ViTEraser-Small~\cite{peng2024viteraser} but incorporates three learnable prompts for multi-task processing, totally containing 108M parameters.
The encoder, decoder, and task prompts comprise 49.6M, 58.6M, and 2,304 parameters, respectively.
During training, UPOCR are initialized using pretrained ViTEraser-Small (with SegMIM pre-training).
The input size is set to $512\times512$.

\noindent\textbf{Other Details.}
\textit{See Sec.~\ref{sec:appendix_implementation_details} of the appendix for implementation details.}

\begin{table*}[t]
    \centering 
    \caption{Comparison with \textbf{specialized} models for text removal on the SCUT-EnsText dataset.
     For a fair comparison, MTRNet++ uses empty coarse masks and GaRNet uses text masks from pretrained CRAFT~\cite{baek2019character} instead of leveraging GT text masks.}
    \label{tab:text_removal}
    \resizebox{1.65\columnwidth}{!}{
        \begin{tabular}{rcccccccccc}
        \toprule 
        \multirow{2}*{Method} & \multicolumn{7}{c}{Image-Eval} & \multicolumn{3}{c}{Detection-Eval} \\
        \cmidrule(r){2-8} \cmidrule(r){9-11}
        & PSNR$\uparrow$ & MSSIM$\uparrow$ & MSE$\downarrow$ & AGE$\downarrow$ & pEPs$\downarrow$ & pCEPs$\downarrow$ & FID$\downarrow$ & R$\downarrow$ & P$\downarrow$ & F$\downarrow$ \\
        \midrule 
        Original & - & - & - & - & - & - & - & 69.5 & 79.4 & 74.1 \\ 
        Pix2pix~\cite{isola2017image} & 26.70 & 88.56 & 0.37 & 6.09 & 0.0480 & 0.0227 & 46.88 & 35.4 & 69.7 & 47.0 \\
        STE~\cite{nakamura2017scene} & 25.47 & 90.14 & 0.47 & 6.01 & 0.0533 & 0.0296 & 43.39 & 5.9 & 40.9 & 10.2 \\
        EnsNet~\cite{zhang2019ensnet} & 29.54 & 92.74 & 0.24 & 4.16 & 0.0307 & 0.0136 & 32.71 & 32.8 & 68.7 & 44.4 \\
        MTRNet++~\cite{tursun2020mtrnet++} & 29.63 & 93.71 & 0.28 & 3.51 & 0.0305 & 0.0168 & 35.68 & 15.1 & 63.8 & 24.4 \\
        EraseNet~\cite{liu2020erasenet} & 32.30 & 95.42 & 0.15 & 3.02 & 0.0160 & 0.0090 & 19.27 & 4.6 & 53.2 & 8.5 \\
        SSTE~\cite{tang2021stroke} & 35.34 & 96.24 & 0.09 & - & - & - & - & 3.6 & - & - \\ 
        PSSTRNet~\cite{lyu2022psstrnet} & 34.65 & 96.75 & 0.14 & \underline{1.72} & 0.0135 & 0.0074 & - & 5.1 & 47.7 & 9.3 \\
        CTRNet~\cite{liu2022don} & 35.20 & 97.36 & 0.09 & 2.20 & 0.0106 & 0.0068 & 13.99 & 1.4 & 38.4 & 2.7 \\
        GaRNet~\cite{lee2022surprisingly} & 35.45 & 97.14 & 0.08 & 1.90 & 0.0105 & 0.0062 & 15.50 & 1.6 & 42.0 & 3.0 \\
        MBE~\cite{hou2022multi} & 35.03 & 97.31 & - & 2.06 & 0.0128 & 0.0088 & - & - & - & - \\
        PEN~\cite{du2022progressive} & 35.72 & 96.68 & 0.05 & 1.95 & 0.0071 & 0.0020 & - & 2.1 & \textbf{26.2} & 3.9  \\
        PERT~\cite{wang2023pert} & 33.62 & 97.00 & 0.13 & 2.19 & 0.0135 & 0.0088 & - & 4.1 & 50.5 & 7.6 \\ 
        SAEN~\cite{du2023modeling} & 34.75 & 96.53 & 0.07 & 1.98 & 0.0125 & 0.0073 & - & - & - & - \\
        FETNet~\cite{lyu2023fetnet} & 34.53 & 97.01 & 0.13 & 1.75 & 0.0137 & 0.0080 & - & 5.8 & 51.3 & 10.5 \\
        ViTEraser-Base~\cite{peng2024viteraser} & \underline{37.11} & \underline{97.61} & \underline{0.0474} & \textbf{1.70} & \underline{0.0066} & \underline{0.0035} & \textbf{10.15} & \textbf{0.389} & \underline{29.7} & \textbf{0.768} \\
        \midrule 
        UPOCR (Ours) & \textbf{37.14} & \textbf{97.62} & \textbf{0.0428} & \underline{1.72} & \textbf{0.0064} & \textbf{0.0034} & \underline{10.47} & \underline{0.614} & 36.6 & \underline{1.208} \\
        \bottomrule
    \end{tabular}}
\end{table*}

\subsection{Evaluation Metrics}
\noindent\textbf{Text Removal.}
Following previous methods~\cite{liu2020erasenet,liu2022don}, the evaluation for text removal involves image- and detection-eval metrics.
The image-eval metrics include PSNR, MSSIM, MSE, AGE, pEPs, pCEPs, and FID while the detection-eval metrics are precision (P), recall (R), and f-measure (F) using the pretrained text detector CRAFT~\cite{baek2019character}.
Note that MSSIM and MSE are presented in percent (\%).

\noindent\textbf{Text Segmentation.}
The intersection over union (IoU), P, R, and F of foreground pixels are employed for evaluation on text segmentation~\cite{xu2021rethinking}.

\noindent\textbf{Tampered Text Detection.}
The evaluation metrics include the P, R, F, and IoU of both real and tampered pixels.
Additionally, the mean f-measure (mF) and mean IoU (mIoU) of real and tampered pixels are calculated for overall evaluation following \citet{wang2022detecting}.

\subsection{Ablation Study}

\noindent\textbf{Insertion of Task Prompts.}
The crucial component of the generalist proficiency of UPOCR is the learnable task prompt. 
As specified in Sec.~\ref{sec:method_UA} and Fig.~\ref{fig:method}(b), the task prompt is inserted to the shared feature space between the encoder and decoder, following the idea that the encoder extracts versatile OCR-related features and the decoder decodes task-specific output images from adjusted hidden features.
Nevertheless, the task prompts can also be inserted into the intermediate features of the encoder and decoder, after each encoder block and decoder block. 
In Tab.~\ref{tab:prompt_insertion}, we investigate different inserting positions of task prompts.
If task prompts are supposed to be integrated into the encoder or decoder, they will go through linear layers to match the required dimensions of intermediate features.
The experimental results suggest simply inserting task prompts into the shared feature space can effectively prompt the decoder to generate images for specific tasks, 
Moreover, it also brings extremely small overhead on parameters (2.3K) and computational costs (only an element-wise addition to a $16\times16\times768$ feature map).


\noindent\textbf{Task Collaboration.}
In Tab.~\ref{tab:task_collaboration}, we conduct the experiments with different task compositions.
Based on these results, We come to the following insights.
(1) \textit{The task with insufficient samples can easily benefit from the joint training with other tasks.} 
For example, the Tampered-IC13~\cite{wang2022detecting} dataset used for tampered text detection contains only 229 training samples, which is inadequate to train a network with 108M parameters.
Therefore, the performance of tampered text detection is consistently improved because the fundamental text localization capacity can be boosted by other tasks.
(2) \textit{The joint training of tasks leads to mutual collaboration if with similar targets but negative effect if with exclusive targets.}
For instance, text removal aims at erasing the texts while text segmentation needs to highlight them at a fine-grained stroke level.
Therefore, additional text removal task always leads to downgraded text segmentation performance.
However, comparing the 2\textit{nd} and 5\textit{th} rows, tampered text detection helps improve the accuracy of text segmentation, because the former provides auxiliary polygon-level text location supervision.
Due to the large model capacity, the final model can achieve a good balance of three tasks.

\begin{table}[t]
    \centering 
    \caption{Comparison with \textbf{specialized} models for text segmentation on the TextSeg dataset.
    The performances of U-Net and SegFormer are cited from \cite{ren2022looking} and \cite{yu2023scene}, respectively.
    }
    \label{tab:text_seg}
    \resizebox{1\columnwidth}{!}{
        \begin{tabular}{rcccc}
            \toprule 
            Method & fgIoU$\uparrow$ & P$\uparrow$ & R$\uparrow$ & F$\uparrow$ \\
            \midrule
            U-Net~\cite{ronneberger2015u} & - & 89.00 & 77.40 & 82.80 \\
            DeepLabV3+~\cite{chen2018encoder} & 84.07 & - & - & 91.40 \\
            HRNetv2-W48~\cite{wang2020deep} & 85.03 & - & - & 91.40 \\
            HRNetv2-W48+OCR~\cite{wang2020deep} & 85.98 & - & - & 91.80 \\
            TexRNet+DeepLabV3+~\cite{xu2021rethinking} & 86.06 & - & - & 92.10 \\
            TexRNet+HRNetv2-W48~\cite{xu2021rethinking} & 86.84 & - & - & 92.40 \\
            SegFormer~\cite{xie2021segformer} & 84.59 & - & - & 91.60 \\
            ARM-Net~\cite{ren2022looking} & - & \underline{92.80} & \underline{92.60} & 92.70 \\
            TFT~\cite{yu2023scene} & \underline{87.11} & - & - & \underline{93.10} \\
            \midrule 
            UPOCR (Ours) & \textbf{88.76} & \textbf{94.55} & \textbf{93.55} & \textbf{94.04} \\
            \bottomrule
        \end{tabular}
    }
\end{table}

\begin{table*}[!h]
    \centering 
    \caption{Comparsion with \textbf{specialized} models for tampered text detection on the Tampered-IC13 dataset.
    CAT-Net can only perform binary classification between tampered texts and backgrounds.
    The \textbf{\textcolor{blue}{mIoU}} and \textbf{\textcolor{blue}{mF}} metrics indicate the overall effectiveness on this task.}
    \label{tab:ttd}
    \resizebox{1.9\columnwidth}{!}{
        \begin{tabular}{rcccccccccc}
            \toprule 
            \multirow{2}*{Method} & \multicolumn{4}{c}{Real Text} & \multicolumn{4}{c}{Tampered Text} & \textbf{\textcolor{blue}{\multirow{2}*{mIoU$\uparrow$}}} & \textbf{\textcolor{blue}{\multirow{2}*{mF$\uparrow$}}} \\
            \cmidrule(r){2-5} \cmidrule(r){6-9}
            & IoU$\uparrow$ & P$\uparrow$ & R$\uparrow$ & F$\uparrow$ & IoU$\uparrow$ & P$\uparrow$ & R$\uparrow$ & F$\uparrow$ & & \\
            \midrule 
            \multicolumn{11}{c}{Detection-based Methods} \\
            \hline
            S3R~\cite{wang2022detecting}+ContourNet~\cite{wang2020contournet} & - & 77.88 & 54.80 & 64.33 & - & 86.68 & \textbf{91.45} & \underline{88.99} & - & 76.66 \\
            ViTEraser~\cite{peng2024viteraser}+ContourNet~\cite{wang2020contournet} & - & 56.84 & \textbf{75.82} & 64.97 & - & \textbf{92.62} & 85.77 & \textbf{89.06} & - & 77.02 \\
            \midrule
            \multicolumn{11}{c}{Segmentation-based Methods} \\
            \hline
            DeepLabV3+~\cite{chen2018encoder} & 48.12 & 79.83 & 54.78 & 64.98 & 72.21 & 89.75 & 78.71 & 83.86 & 60.17 & 74.42 \\
            HRNetv2~\cite{wang2020deep} & 43.26 & 76.35 & 49.95 & 60.39 & 73.12 & 89.98 & 79.60 & 84.47 & 58.19 & 72.43 \\
            Swin-Uper~\cite{liu2021swin} & \underline{61.82} & \underline{87.82} & 67.62 & \underline{76.41} & \underline{77.28} & 89.67 & 84.83 & 87.18 & \underline{69.55} & \underline{81.80} \\
            SegFormer~\cite{xie2021segformer} & 53.22 & 86.39 & 58.09 & 69.47 & \textbf{77.78} & \underline{91.78} & 83.60 & 87.50 & 65.50 & 78.49\\
            BEiT-Uper~\cite{bao2022beit} & 57.07 & 81.23 & 65.74 & 72.67 & 70.88 & 82.27 & 83.66 & 82.96 & 63.98 & 77.82 \\
            CAT-Net~\cite{kwon2022learning} & - & - & - & - & 28.31 & 31.45 & 73.91 & 44.13 & - & - \\
            \midrule
            UPOCR (Ours) & \textbf{71.80} & \textbf{93.31} & \underline{75.70} & \textbf{83.59} & 71.62 & 79.76 & \underline{87.53} & 83.46 & \textbf{71.71} & \textbf{83.53} \\
            \bottomrule
        \end{tabular}
    }
\end{table*}

\begin{figure*}[!h]
    \centering
    \includegraphics[width=1.9\columnwidth]{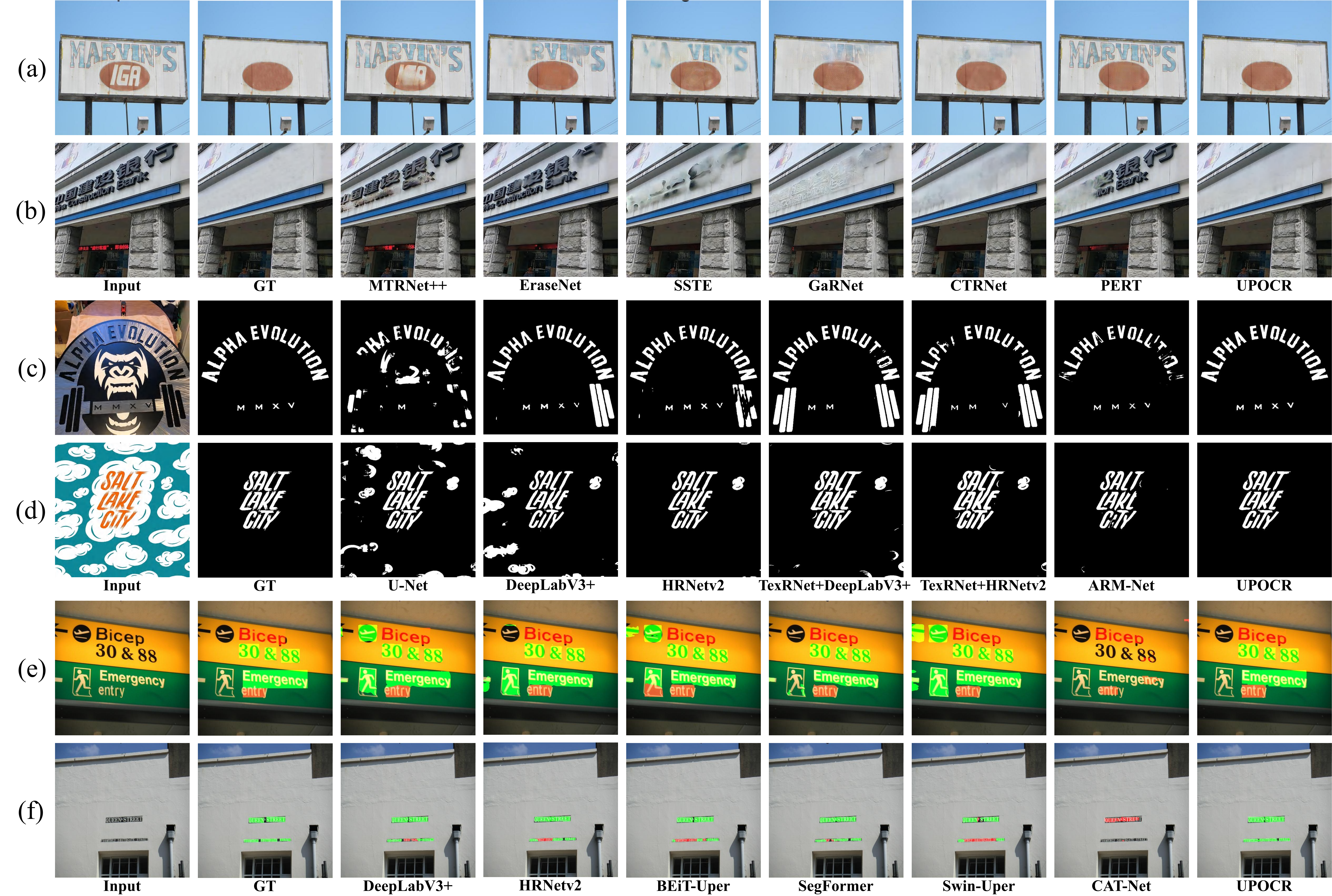}
    \caption{Qualitative comparison of UPOCR and existing \textbf{specialized} models on (a)-(b) text removal, (c)-(d) text segmentation, and (e)-(f) tampered text detection (red: tampered, green: real). 
    Zoom in for a better view.}
    \label{fig:vis}
\end{figure*}

\begin{table*}
    \centering
    \caption{Comparison with \textbf{generalist} models on pixel-level OCR tasks.}
    \label{tab:painter}
    \resizebox{1.9\columnwidth}{!}{
        \begin{tabular}{rccccccccccccc}
            \toprule
            \multirow{2}*{Method} & \multicolumn{7}{c}{Text Removal} & \multicolumn{4}{c}{Text Segmentation} & \multicolumn{2}{c}{Tampered Text Det.} \\
            \cmidrule(r){2-8} \cmidrule(r){9-12} \cmidrule(r){13-14}
            & PSNR$\uparrow$ & MSSIM$\uparrow$ & MSE$\downarrow$ & AGE$\downarrow$ & pEPs$\downarrow$ & pCEPs$\downarrow$ & FID$\downarrow$ & fgIoU$\uparrow$ & P$\uparrow$ & R$\uparrow$ & F$\uparrow$ & mIoU$\uparrow$ & mF$\uparrow$ \\
            \midrule 
            Painter~\cite{wang2023images} & 27.13 & 91.67 & 0.2942 & 8.68 & 0.0898 & 0.0425 & 21.90 & 86.36 & 93.40 & 91.97 & 92.68 & 69.26 & 81.83 \\
            UPOCR (Ours) & \textbf{37.14} & \textbf{97.62} & \textbf{0.0428} & \textbf{1.72} & \textbf{0.0064} & \textbf{0.0034} & \textbf{10.47} & \textbf{88.76} & \textbf{94.55} & \textbf{93.55} & \textbf{94.04} & \textbf{71.71} & \textbf{83.53} \\
            \bottomrule
        \end{tabular}
    }
\end{table*}

\subsection{Comparison with Specialized Models}
The comparisons of UPOCR with existing specialized methods for text removal, text segmentation, and tampered text detection are presented in Tabs.~\ref{tab:text_removal}, \ref{tab:text_seg}, and \ref{tab:ttd}, respectively.
Furthermore, the visualization results on three tasks are shown in Fig.~\ref{fig:vis}.
Without bells and whistles, the generalist UPOCR with shared parameters can simultaneously outperform existing specialized models for individual tasks.
(1) \textbf{\textit{Text Removal}}: 
The UPOCR eschews the complicated text localization modules, external text detectors, and multi-step refinements.
Furthermore, UPOCR discards the GAN-based training strategy and mask branch of ViTEraser during training.
Following a concise pipeline, UPOCR achieves state-of-the-art image-eval performance on SCUT-EnsText~\cite{liu2020erasenet}, outperforming ViTEraser-Base~\cite{peng2024viteraser} with a lighter-weight ViTEraser-Small architecture.
As for detection-eval metrics, the 0.614\% recall of UPOCR is comparable to ViTEraser, demonstrating nearly all texts are exhaustively erased.
It is worth noting that compared with earlier approaches, UPOCR significantly outperforms them in terms of both image- and detection-eval metrics.
(2) \textbf{\textit{Text Segmentation.}}
Although previous methods utilize attentive modules for refinements, text detectors for coarse text localization, and text recognizers for semantic supervision, UPOCR significantly outperforms them on TextSeg~\cite{xu2021rethinking} using a single encoder-decoder without extra modules and annotations.
(3) \textbf{\textit{Tampered Text Detection.}}
The UPOCR achieves the best performance of 71.71\% mIoU and 83.53\% mF on Tampered-IC13~\cite{wang2022detecting}, eliminating the need for frequency domain fusion~\cite{qu2023towards,kwon2022learning,wang8tampered} and adaptation based on text detectors~\cite{wang2022detecting}.
Note that the segmentation-based methods are reimplemented using MMSegmentation\footnote{https://github.com/open-mmlab/mmsegmentation} and CAT-Net's official codes\footnote{https://github.com/mjkwon2021/CAT-Net}.

\begin{figure}
    \centering
    \includegraphics[width=0.9\columnwidth]{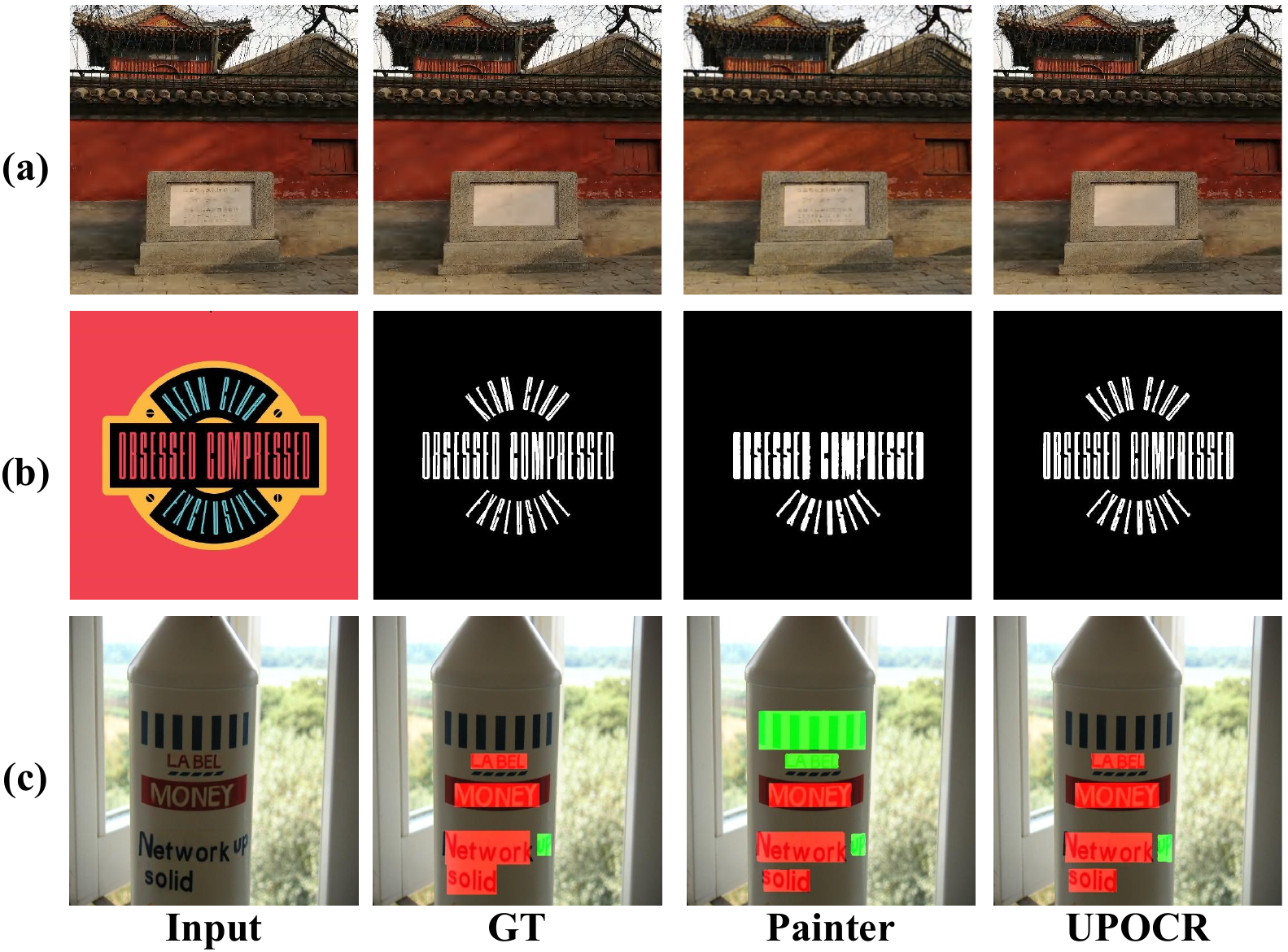}
    \caption{Qualitative comparison of UPOCR and Painter on (a) text removal, (b) text segmentation, and (c) tampered text detection (red: tampered, green: real). 
    Zoom in for a better view.}
    \label{fig:painter}
\end{figure}

\subsection{Comparison with Generalist Models}
\label{sec:exp_generalist}

To demonstrate the effectiveness of UPOCR over existing generalist models, we re-train Painter~\cite{wang2023images} using the official implementation\footnote{https://github.com/baaivision/Painter} and the same dataset as ours.
The performances of UPOCR and Painter are listed in Tab.~\ref{tab:painter} while the visualizations are illustrated in Fig.~\ref{fig:painter}.
Because Painter learns the task target from an example input-output pair, it may hardly grasp the inconspicuous correlation such as tiny text erasing (Fig.~\ref{fig:painter}(a)) and a strong ability to distinguish texts from text-like patterns (Fig.~\ref{fig:painter}(c)).
Moreover, as the image is predicted through inpainting, the output cannot guarantee consistent colors as inputs, \textit{e.g.}, red vs. orange walls in Fig.~\ref{fig:painter}(a).
Therefore, UPOCR surpasses Painter on all three tasks, especially the text removal task.

\begin{figure}[t]
    \centering
    \includegraphics[width=1\columnwidth]{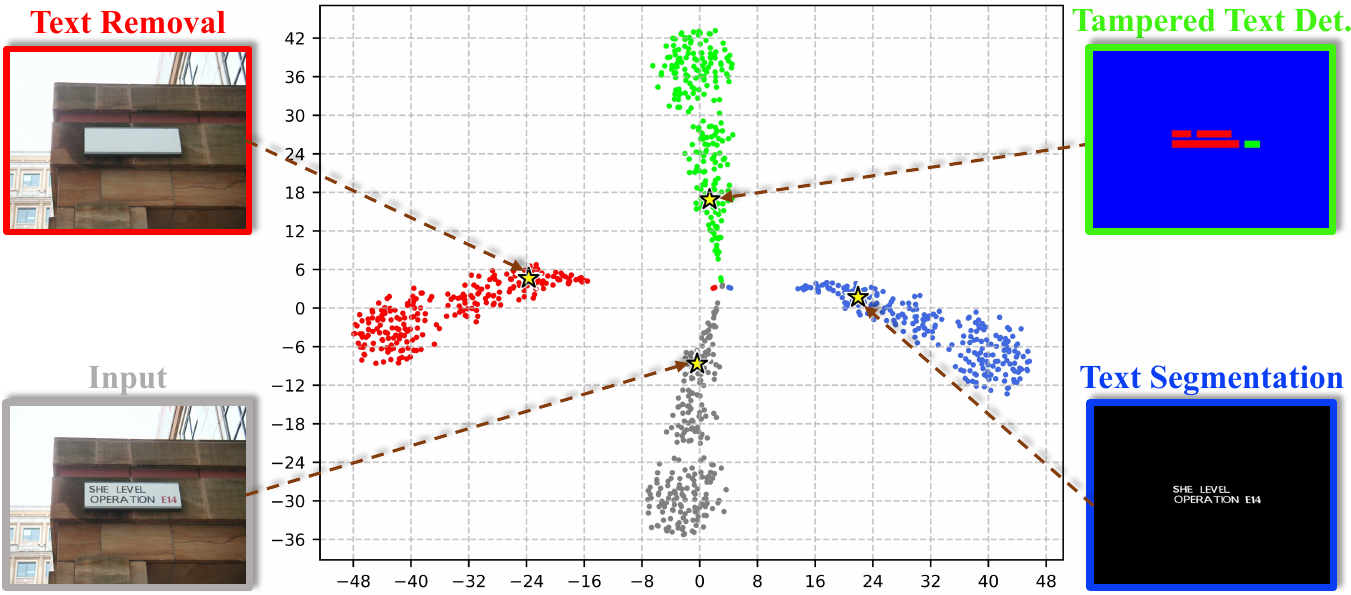}
    \caption{The t-SNE visualization of general features (gray circles) extracted by the encoder and task-specific features integrated with task prompts. 
             Red, green and blue circles are for text removal, tampered text detection, and text segmentation, respectively.
             The example input image and three-task outputs are visualized along with corresponding features (yellow stars).}
    \label{fig:task_prompt}
\end{figure}

\subsection{Interpretability of Task Prompts}
To investigate the mechanism of task prompts, we visualize the features before and after integrated with task prompts in Fig.~\ref{fig:task_prompt}.
Specifically, we perform all three tasks on the 233 testing samples of Tampered-IC13 using corresponding task prompts.
We can observe that there are clear boundaries between the general features (gray circles) extracted by the encoder and the task-specific features (red, green, and blue circles), indicating the proposed task prompts can effectively adapt the general features into task-specific regions to generate output images for individual tasks.

\section{Conclusion}
In this paper, we propose UPOCR, a first-of-its-kind simple-yet-effective unified pixel-level OCR interface.
To acquire generalist capability, UPOCR unifies the paradigm, architecture, and training strategy of diverse pixel-level OCR tasks.
Specifically, existing divergent paradigms are unified as RGB image to RGB image transformation.
To implement this paradigm, UPOCR uniformly adopts a ViT-based encoder-decoder with learnable task prompts to handle various tasks.
During training, the strategy is unified to minimize the discrepancy between the predicted and ground-truth images at pixel and feature spaces.
Extensive experiments are conducted on text removal, text segmentation, and tampered text detection to verify the generalist proficiency of UPOCR.
The experimental results demonstrate that UPOCR simultaneously achieves state-of-the-art performance with shared parameters, significantly surpassing specialized OCR models.
Comprehensive ablation studies and visual analyses are also presented to provide in-depth insights. 
We believe this work could be extended to broader tasks and spark more research on generalist OCR models.

\section*{Acknowledgement}
This research is supported in part by National Natural Science Foundation of China (Grant No.: 62441604, 61936003).

\section*{Impact Statement}
This paper presents work whose goal is to advance the fields of optical character recognition and machine learning. There are many potential societal consequences of our work, none which we feel must be specifically highlighted here.

\bibliography{main}
\bibliographystyle{icml2024}

\newpage
\appendix
\onecolumn
\section{Implementation Details}
\label{sec:appendix_implementation_details}
In this section, we supplement every detail to implement the proposed UPOCR, which may not be thoroughly specified in the main paper due to the page limit.

\subsection{Network Architecture}
\label{sec:appendix_network_architecture}
As described in Secs.~\ref{sec:method_UA} and \ref{sec:exp_set} of the main paper, the UPOCR employs a vision Transformer (ViT)-based encoder-decoder with learnable task prompts.
The detailed network architecture of UPOCR is presented in Tab.~\ref{tab:arch} where the notations are defined as follows.
\begin{itemize}
    \item $D_{i}^{enc}$: The downsampling ratio of the patch embedding layer in the \textit{i}-th block of the encoder.
    \item $E_{i}^{enc}$: The output dimension of the patch embedding layer in the \textit{i}-th block of the encoder.
    \item $W_{i}^{enc}$: The window size of the Swinv2~\cite{liu2022swin} blocks in the \textit{i}-th block of the encoder.
    \item $H_{i}^{enc}$: The number of heads of the Swinv2 blocks in the \textit{i}-th block of the encoder.
    \item $C_{i}^{enc}$: The feature dimension of the Swinv2 blocks in the \textit{i}-th block of the encoder.
    \item $N^{tp}$: The number of task prompts which is equal to the number of tasks that the model is supposed to simultaneously deal with.
    \item $C^{tp}$: The dimension of task prompts.
    \item $U_{i}^{dec}$: The upsampling ratio of the patch splitting layer in the \textit{i}-th block of the decoder.
    \item $E_{i}^{dec}$: The output dimension of the patch splitting layer in the \textit{i}-th block of the decoder.
    \item $W_{i}^{dec}$: The window size of the Swinv2 blocks in the \textit{i}-th block of the decoder.
    \item $H_{i}^{dec}$: The number of heads of the Swinv2 blocks in the \textit{i}-th block of the decoder.
    \item $C_{i}^{dec}$: The feature dimension of the Swinv2 blocks in the \textit{i}-th block of the decoder.
\end{itemize}

\begin{table}[t]
    \centering 
    \caption{Detailed network architecture of the proposed UPOCR.}
    \label{tab:arch}
    \resizebox{0.5\columnwidth}{!}{
    \begin{tabular}{c|c|c|c|c}
    \hline
    & Block & Output Size & Layer Name & Details \\
    \hline 
    \hline
    \multirow{8}{*}[-5em]{\rotatebox{90}{Encoder}} & \multirow{2}{*}[-1.25em]{Block 1} & \multirow{2}{*}[-1.25em]{\Large $\frac{H}{4} \times \frac{W}{4}$} & Patch Embedding & $D_1^{enc} = 4$, $E_1^{enc} = 96$ \\
    \cline{4-5}
    & & & Swinv2 block & $\left [ \begin{matrix} W_1^{enc} = 16 \\ H_1^{enc} = 3 \\ C_1^{enc} = 96 \end{matrix} \right ] \times 2$ \\
    \cline{2-5}

    & \multirow{2}{*}[-1.25em]{Block 2} & \multirow{2}{*}[-1.25em]{\Large $\frac{H}{8} \times \frac{W}{8}$} & Patch Embedding & $D_2^{enc} = 2$, $E_2^{enc} = 192$ \\
    \cline{4-5}
    & & & Swinv2 block & $\left [ \begin{matrix} W_2^{enc} = 16 \\ H_2^{enc} = 6 \\ C_2^{enc} = 192 \end{matrix} \right ] \times 2$ \\
    \cline{2-5}

    & \multirow{2}{*}[-1.25em]{Block 3} & \multirow{2}{*}[-1.25em]{\Large $\frac{H}{16} \times \frac{W}{16}$} & Patch Embedding & $D_3^{enc} = 2$, $E_3^{enc} = 384$ \\
    \cline{4-5}
    & & & Swinv2 block & $\left [ \begin{matrix} W_3^{enc} = 16 \\ H_3^{enc} = 12 \\ C_3^{enc} = 384 \end{matrix} \right ] \times 18$ \\
    \cline{2-5}

    & \multirow{2}{*}[-1.25em]{Block 4} & \multirow{2}{*}[-1.25em]{\Large $\frac{H}{32} \times \frac{W}{32}$} & Patch Embedding & $D_4^{enc} = 2$, $E_4^{enc} = 768$ \\
    \cline{4-5}
    & & & Swinv2 block & $\left [ \begin{matrix} W_4^{enc} = 16 \\ H_4^{enc} = 24 \\ C_4^{enc} = 768 \end{matrix} \right ] \times 2$ \\
    \hline
    \hline

    \multicolumn{2}{c|}{Task Prompts} & {\Large $\frac{H}{32} \times \frac{W}{32}$} & \multicolumn{2}{c}{\normalsize $N^{tp} = 3, C^{tp} = 768$}\\

    \hline
    \hline
    \multirow{8}{*}[-6em]{\rotatebox{90}{Decoder}} & \multirow{2}{*}{Block 1} & \multirow{2}{*}{\Large $\frac{H}{16} \times \frac{W}{16}$} & Swinv2 block & $\left [ \begin{matrix} W_1^{dec} = 8 \\ H_1^{dec} = 24 \\ C_1^{dec} = 768 \end{matrix} \right ] \times 2$ \\
    \cline{4-5}
    & & & Patch Splitting & $U_1^{dec} = 2$, $E_1^{dec} = 384$ \\
    \cline{2-5}

    & \multirow{2}{*}{Block 2} & \multirow{2}{*}{\Large $\frac{H}{8} \times \frac{W}{8}$} & Swinv2 block & $\left [ \begin{matrix} W_2^{dec} = 8 \\ H_2^{dec} = 12 \\ C_2^{dec} = 384 \end{matrix} \right ] \times 18$ \\
    \cline{4-5}
    & & & Patch Splitting & $U_2^{dec} = 2$, $E_2^{dec} = 192$ \\
    \cline{2-5}

    & \multirow{2}{*}{Block 3} & \multirow{2}{*}{\Large $\frac{H}{4} \times \frac{W}{4}$} & Swinv2 block & $\left [ \begin{matrix} W_3^{dec} = 8 \\ H_3^{dec} = 6 \\ C_3^{dec} = 192 \end{matrix} \right ] \times 2$ \\
    \cline{4-5}
    & & & Patch Splitting & $U_3^{dec} = 2$, $E_3^{dec} = 96$ \\
    \cline{2-5}

    & \multirow{2}{*}{Block 4} & \multirow{2}{*}{\Large $\frac{H}{2} \times \frac{W}{2}$} & Swinv2 block & $\left [ \begin{matrix} W_4^{dec} = 8 \\ H_4^{dec} = 3 \\ C_4^{dec} = 96 \end{matrix} \right ] \times 2$ \\
    \cline{4-5}
    & & & Patch Splitting & $U_4^{dec} = 2$, $E_4^{dec} = 48$ \\
    \cline{2-5}

    & \multirow{2}{*}{Block 5} & \multirow{2}{*}{$H \times W$} & Swinv2 block & $\left [ \begin{matrix} W_5^{dec} = 8 \\ H_5^{dec} = 2 \\ C_5^{dec} = 48 \end{matrix} \right ] \times 2$ \\
    \cline{4-5}
    & & & Patch Splitting & $U_5^{dec} = 2$, $E_5^{dec} = 24$ \\
    \hline
    \end{tabular}}
\end{table}

As shown in Tab.~\ref{tab:arch}, supposing the shape of the input RGB image is $H \times W \times 3$, the encoder hierarchically produces features $\{f_{i}^{enc} \in \mathbb{R}^{\frac{H}{2^{(i+1)}} \times \frac{W}{2^{(i+1)}} \times C_{i}^{enc}}\}_{i=1}^{4}$.
Then the learnable prompt of the target task is integrated into the final feature $f_{4}^{enc}$ of the encoder, yielding a task-specific feature for decoding.
Subsequently, the decoder hierarchically generates features $\{f_{i}^{dec} \in \mathbb{R}^{\frac{H}{2^{(5-i)}} \times \frac{W}{2^{(5-i)}} \times C_{i}^{dec}}\}_{i=1}^{5}$ based on the task-specific feature.
Finally, the output image is predicted based on feature $f_{5}^{dec}$ through a $3 \times 3$ convolution.

\noindent\textbf{Task Prompt Insertion.}
The task prompts are formulated as a set of learnable embeddings $f^{tp} \in \mathbb{R}^{N^{tp} \times C^{tp}}$, where $N^{tp} = 3$ and $C^{tp} = 768$ are the number of tasks and embedding dimension, respectively.
To perform the \textit{i}-th task, the correspond prompt $f_{i}^{tp} \in \mathbb{R}^{1 \times 768}$ is first repeated by $\frac{H}{32} \times \frac{W}{32}$ times, yielding a feature $\hat{f}_{i}^{tp} \in \mathbb{R}^{\frac{H}{32} \times \frac{W}{32} \times 768}$.
Then the $\hat{f}_{i}^{tp}$ is element-wise added to the final feature $f_{4}^{enc} \in \mathbb{R}^{\frac{H}{32} \times \frac{W}{32} \times 768}$ of the encoder, pushing the general representations towards task-specific regions.

\noindent\textbf{Patch Embedding.}
Given an input feature map or image $f_{in} \in \mathbb{R}^{h \times w \times c_{in}}$, a patch embedding layer with a downsampling ratio of $r$ and an output dimension of $c_{out}$ first flattens each $r \times r$ patch, yield an intermediate feature $f' \in \mathbb{R}^{\frac{h}{r} \times \frac{w}{r} \times r^2 c_{in}}$. 
Then a linear layer is adopted to transform the dimension of feature $f'$, producing the output feature $f_{out} \in \mathbb{R}^{\frac{h}{r} \times \frac{w}{r} \times c_{out}}$.

\noindent\textbf{Patch Splitting.}
Supposing the input feature is $f_{in} \in \mathbb{R}^{h \times w \times c_{in}}$, a patch splitting layer with $r$ upsampling ratio and $c_{out}$ output dimension first generates an intermediate feature $f' \in \mathbb{R}^{rh \times rw \times \frac{c_{in}}{r^2}}$ by decomposing each pixel of the input feature into a $r \times r$ patch.
Subsequently, the feature $f'$ goes through a linear layer to transform the feature dimension, yielding the output feature $f_{out} \in \mathbb{R}^{rh \times rw \times c_{out}}$.

\noindent\textbf{Lateral Connection.}
As shown in Fig.~\ref{fig:method}(b) of the main paper, lateral connections are built between the encoder and decoder blocks to shortcut the transmission of fine-grained representations.
Specifically, the encoder features $\{f_{i}^{enc}\}_{i=1}^{3}$ are laterally connected to the decoder features $\{f_{4-i}^{dec}\}_{i=1}^{3}$.
As for the architecture, if the feature $f_{1} \in \mathbb{R}^{h \times w \times c}$ is connected to feature $f_{2} \in \mathbb{R}^{h \times w \times c}$, the lateral connection processes the feature $f_1$ sequentially using a $1 \times 1$ convolution with $c$ channels for non-linear transformation, two $3 \times 3$ convolutions with $2c$ channels for expanding, and a $1 \times 1$ convolution with $c$ channels for shrinking, following EraseNet~\cite{liu2020erasenet}.
Then the resulting feature is element-wise added to feature $f_2$.

\subsection{Loss Function}
\noindent\textbf{Pixel Space.}
As described in Sec.~\ref{sec:unified_training_strategy} and defined in Eq.~(\ref{equ:pixel_loss}) of the main paper, the pixel loss is formulated as the weighted sum of L1 distances between multi-scale output images $\mathbb{I}_{out} = \{I_{out}^{\frac{1}{4}}, I_{out}^{\frac{1}{2}}, I_{out}\}$ and ground-truth (GT) images $\mathbb{I}_{gt} = \{I_{gt}^{\frac{1}{4}}, I_{gt}^{\frac{1}{2}}, I_{gt}\}$.

In practical implementation, for samples of the text removal task, the GT text box mask $M_{gt}$ is incorporated to focus more on the discrepancy in text regions, following most existing text removal methods~\cite{liu2020erasenet,zhang2019ensnet,liu2022don}.
Specifically, the $M_{gt} \in \mathbb{R}^{H \times W}$ is a binary mask at the bounding-box level where 1 and 0 values indicate text and non-text pixels, respectively.
Similar to $\mathbb{I}_{gt}$, we also resize $M_{gt}$ to multiple scales denoted as $\mathbb{M}_{gt} = \{M_{gt}^{\frac{1}{4}}, M_{gt}^{\frac{1}{2}}, M_{gt}\}$.
Then the pixel loss of text removal samples is calculated as
\begin{align}
    L_{pix}^{tr} = &\sum\nolimits_{i=1}^{3} \alpha_i || (\mathbb{I}_{out(i)} - \mathbb{I}_{gt(i)}) \odot \mathbb{M}_{gt(i)} ||_1 \notag \\ 
                   & + \beta_i || (\mathbb{I}_{out(i)} - \mathbb{I}_{gt(i)}) \odot (1 - \mathbb{M}_{gt(i)}) ||_1,
\end{align}
where $\alpha$ and $\beta$ are empircally set to $\{5, 6, 10\}$ (as specified in Sec.~\ref{sec:unified_training_strategy} of the main paper) and $\{0.8, 1, 2\}$, respectively.

As for the text segmentation and tampered text detection tasks, smooth L1 loss functions are employed in Eq.~(\textcolor{blue}{1}) of the main paper to replace the standard L1 distance, because it is not required to precisely align the pixel values of output and GT images but critical to penalize the outliers for these two segmentation-oriented tasks.

\noindent\textbf{Feature Space.}
As specified in Sec.~\ref{sec:unified_training_strategy} and Eq.~(\ref{equ:feature_loss}) of the main paper, a feature loss $L_{feat}$ containing perceptual loss $L_{per}$ and style loss $L_{sty}$ is adopted to ensure the visual plausibility of generated images.
Concretely, the losses $L_{per}$ and $L_{sty}$ are formulated as 
\begin{align}
I_{out}^{*} = & I_{out} \odot M_{gt} + I_{in} \odot (1 - M_{gt}), \\
L_{per} = & \sum\nolimits_{i=1}^{3} || \Phi_i(I_{out}) - \Phi_i(I_{gt}) ||_1 \notag \\
          & + || \Phi(I_{out}^{*}) - \Phi_i(I_{gt}) ||_1, \\
L_{sty} = & \sum\nolimits_{i=1}^{3} || Gram(\Phi_i(I_{out})) - Gram(\Phi_i(I_{gt})) ||_1 \notag \\
          & + || Gram(\Phi_i(I_{out}^{*})) - Gram(\Phi_i(I_{gt})) ||_1,
\end{align}
where $I_{in}$ is the input image and $Gram(\cdot)$ calculates the Gram matrix of the input feature map.
Moreover, the $\Phi_i(x)$ represents the feature map produced by the \textit{i}-th pooling layer of an ImageNet~\cite{deng2009imagenet}-pretrained VGG-16~\cite{Simonyan15} network fed with an input $x$.

\subsection{Dataset Statistics}
As described in Sec.~\ref{sec:exp_set} of the main paper, experiments are conducted using the SCUT-EnsText~\cite{liu2020erasenet}, TextSeg~\cite{xu2021rethinking}, and Tampered-IC13~\cite{wang2022detecting} datasets for the text removal, text segmentation, and tampered text detection tasks, respectively.
The statistics of these datasets are introduced as follows.

\noindent\textbf{SCUT-EnsText} is a real-world scene text removal dataset, comprising 2,749 samples for training and 813 samples for testing.

\noindent\textbf{TextSeg} is a large-scale fine-annotated text segmentation dataset with 4,024 images of scene text and design text.
The training, validating, and testing sets contain 2,646, 340, and 1,038 samples, respectively.

\noindent\textbf{Tampered-IC13} is aimed at tampered scene text detection in the wild.
The dataset is divided into 229 training samples and 233 testing samples.
The annotation includes bounding boxes of real and tampered texts.

\subsection{Training Setting}
\label{sec:train_setting}

The proposed UPOCR is implemented with PyTorch\footnote{https://pytorch.org/}.
During training, the parameters of UPOCR are initialized using the pretrained ViTEraser-Small weights (with SegMIM pre-training)~\cite{peng2024viteraser}.
Subsequently, the model is optimized for 80,000 iterations with a batch size of 48 using an AdamW~\cite{loshchilov2018decoupled} optimizer in a multi-task fashion.
Specifically, each batch consists of 16 samples from SCUT-EnsText~\cite{liu2020erasenet} for text removal, 16 samples from TextSeg~\cite{xu2021rethinking} for text segmentation, and 16 samples from Tampered-IC13~\cite{wang2022detecting} for tampered text detection.
The size of training images is set to $512 \times 512$.
The learning rate is initialized as 0.0005 and linearly decays per 200 iterations, finally reaching 0.00001 at the last 200 iterations.
The training lasts approximately 36 hours using two NVIDIA A100 GPUs with 80GB memory.
\textbf{Note that we perform no task- or benchmark-specific finetuning.}

\begin{table*}[t]
    \centering
    \caption{Upper bound of OFA performance on three pixel-level OCR tasks.
             The performance of the proposed UPOCR is also provided for comparison, which has already significantly surpassed the upper bound of OFA on text removal.
            The \textbf{bold} and \underline{underline} indicate the best and the second best, respectively.
            (Seq. Len.: Sequence Length)}
    \label{tab:upper_bound}
    \resizebox{\columnwidth}{!}{
        \begin{tabular}{ccccccccccccc}
            \toprule
            \multirow{2}*{Image Size} & \multirow{2}*{Seq. Len.} & \multicolumn{7}{c}{Text Removal} & \multicolumn{2}{c}{Text Segmentation} & \multicolumn{2}{c}{Tampered Text Det.}  \\
            \cmidrule(r){3-9} \cmidrule(r){10-11} \cmidrule(r){12-13}
            & & PSNR$\uparrow$ & MSSIM$\uparrow$ & MSE$\downarrow$ & AGE$\downarrow$ & pEPs$\downarrow$ & pCEPs$\downarrow$ & FID$\downarrow$ & fgIoU$\uparrow$ & F$\uparrow$ & mIoU$\uparrow$ & mF$\uparrow$ \\
            \midrule
            $512 \times 512$ & 4096 & \underline{24.94} & \underline{78.95} & \underline{0.4871} & \underline{8.74} & \underline{0.1013} & \underline{0.0125} & \underline{19.81} & \textbf{92.97} & \textbf{96.36} & \textbf{96.23} & \textbf{98.07} \\
            $256 \times 256$ & 1024 & 24.50 & 72.88 & 0.5110 & 9.43 & 0.1227 & 0.0286 & 41.24 & 84.15 & 91.39 & \underline{94.98} & \underline{97.42} \\
            \midrule 
            \multicolumn{2}{c}{UPOCR} & \textbf{37.14} & \textbf{97.62} & \textbf{0.0428} & \textbf{1.72} & \textbf{0.0064} & \textbf{0.0034} & \textbf{10.47} & \underline{88.76} & \underline{94.04} & 71.71 & 83.53 \\
            \bottomrule
        \end{tabular}}
\end{table*}

\subsection{Inference}

In this section, we detail the inference procedure mentioned in Sec.~\ref{sec:method_paradigm} of the main paper.
For instance, given an input image $I_{in} \in \mathbb{R}^{H \times W \times 3}$, the UPOCR produces an output image $I_{out} \in \mathbb{R}^{H \times W \times 3}$.
In the following, we introduce how to obtain the formatted prediction for individual tasks from the $I_{out}$ in detail.

\noindent\textbf{Text Removal.}
The output image $I_{out}$ is exactly the text-erased image without additional processing.

\noindent\textbf{Text Segmentation.}
As we define the RGB values of foreground and background pixels as $(255, 255, 255)$ and $(0, 0, 0)$, the text-stroke pixels are determined by setting a threshold of the distance between corresponding RGB values and $(255, 255, 255)$.
To achieve this efficiently, we first normalize $I_{out}$ to the range of $[0, 1]$ and then average the three channels, yielding $\hat{I}_{out} \in \mathbb{R}^{H \times W}$.
Then the pixel at $(i, j)$ position is identified as the text stroke if $\hat{I}_{out}^{(i, j)} > 0.4$ and the background otherwise.

\noindent\textbf{Tampered Text Detection.}
As described in Sec.~\ref{sec:method_paradigm} of the main paper, we compare the distance of the generated RGB values to $(255, 0, 0)$ (red for tamper texts), $(0, 255, 0)$ (green for real texts), and $(0, 0, 255)$ (blue for backgrounds) to determine per-pixel categories.
In practical implementation, it is equivalent to finding the color with the maximum value in the RGB triplet and assigning the corresponding category to the pixel, getting rid of the complex calculation of distances.

\section{Further Comparison with Generalist Model}
In Sec.~\ref{sec:exp_generalist} and Tab.~\ref{tab:painter} of the main paper, we compare the proposed UPOCR with the cutting-edge generalist model (\textit{i.e.}, Painter~\cite{wang2023images}) that is based on image-to-image paradigms.
However, existing generalist models~\cite{wang2022ofa,lu2022unified} with sequence-to-sequence paradigms are also able to perform image-to-image translation.
For instance, given an input image, OFA~\cite{wang2022ofa} produces a sequence composed of discrete tokens from the VQGAN~\cite{esser2021taming} codebook.
After that, the output image is reconstructed from the generated sequence through the decoder of VQGAN.
In this way, the OFA accomplishes the image-to-image transformation in a sequence-to-sequence manner.
Therefore, based on the OFA model, we further conduct experiments and in-depth analyses to verify the effectiveness of sequence-to-sequence generalist models on pixel-level OCR tasks.

\begin{figure}[t]
    \centering 
    \includegraphics[width=0.6\columnwidth]{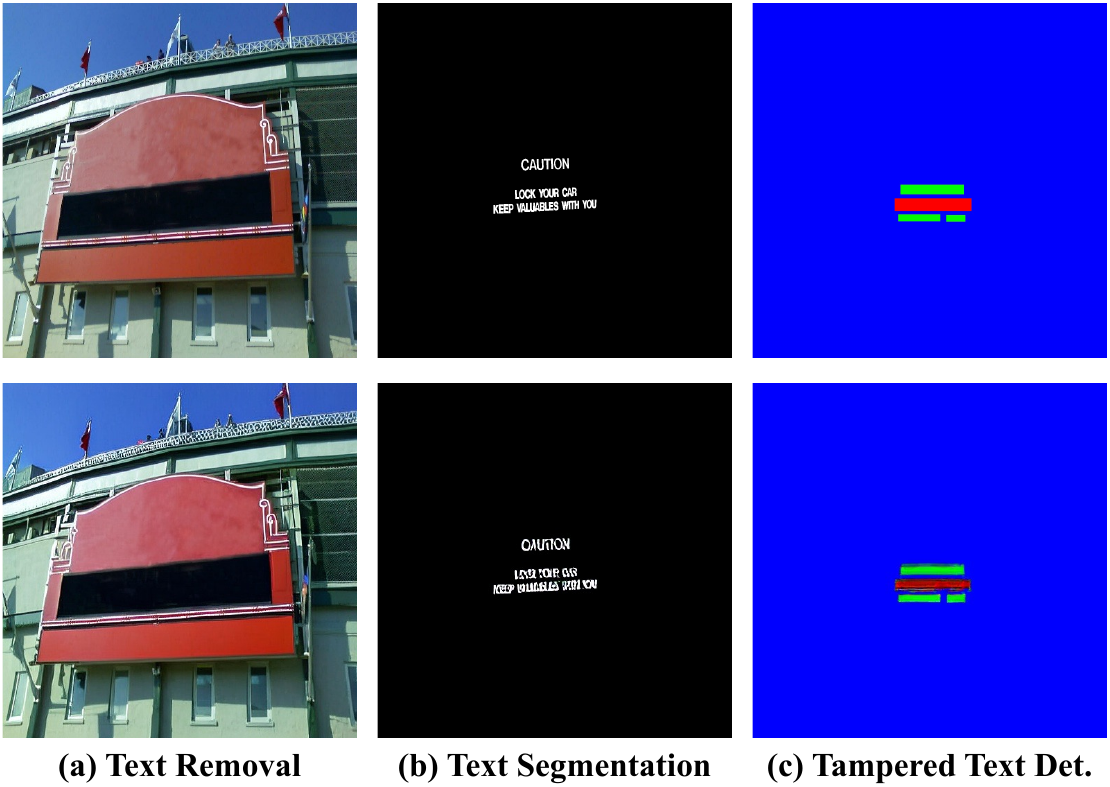}
    \caption{Visualization of GT images (top) and corresponding reconstructed results (bottom) by VQGAN.
    The image size is $512 \times 512$.
    Zoom in for a better view.}
    \label{fig:vis_vqgan}
\end{figure}

\noindent\textbf{Upper Bound of OFA Performance.}
As described above, OFA relies on the VQGAN decoder to reconstruct the output image.
However, there has already been information loss in the encoding and decoding process of VQGAN, limiting the upper bound of OFA performance.
To quantify the information loss, we use VQGAN to encode the GT image $I_{gt}$ into a sequence $S_{gt}$ comprising discrete tokens from the codebook and then directly decode an image $I_{dec}$ from the $S_{gt}$.
Because the OFA is optimized to predict the sequence $S_{gt}$ from the input image, the metrics computed using $I_{dec}$ are the upper bound of OFA performance.

In Tab.~\ref{tab:upper_bound}, we present the upper bounds with two sizes of GT images, including $256 \times 256$ and $512 \times 512$.
Using the VQGAN with a factor of 8 adopted by the OFA, they are encoded into 1024- and 4096-token sequences, respectively.
The metrics listed in Tab.~\ref{tab:upper_bound} demonstrate that the encoding and decoding procedure based on VQGAN itself has caused severe information loss. Especially for the text removal task, the proposed UPOCR has already significantly outperformed the upper bound of OFA.
The reason may be that the decoder of VQGAN cannot reconstruct the details and complex patterns and also struggles to guarantee color consistency as illustrated in Fig.~\ref{fig:vis_vqgan}.

\begin{figure*}
    \centering 
    \includegraphics[width=\columnwidth]{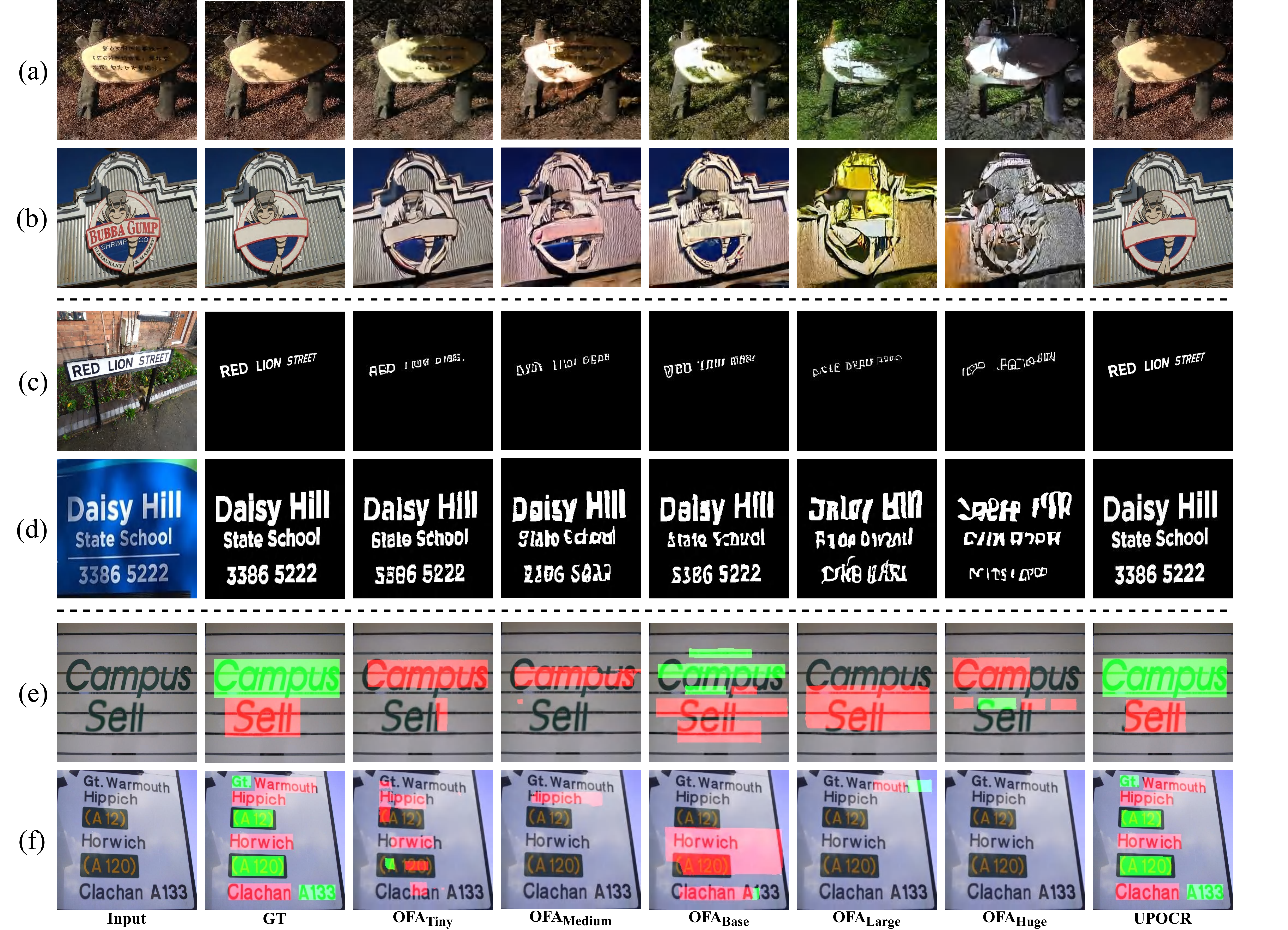}
    \caption{Qualitative comparison of different scales of OFA with UPOCR on (a)-(b) text removal, (c)-(d) text segmentation, and (e)-(f) tampered text detection.
    Zoom in for a better view.}
    \label{fig:ofa}
\end{figure*}

\noindent\textbf{Comparison with OFA.}
Because of the overwhelming computational costs required to train an OFA with a sequence length of 4096, we set the size of input and output images to $256 \times 256$ following the original configuration of OFA.
Tab.~\ref{tab:ofa} presents the comparison between UPOCR and different scales of OFA.
Moreover, the visualizations are shown in Fig.~\ref{fig:ofa}.
Limited by the intrinsic sequence-to-sequence mechanism, OFA exhibits significantly inferior performance on all three pixel-level OCR tasks that require strong image-to-image translation capacity.
\textbf{(1) Text Removal:} As the model size increases, the OFA tends to cause larger color deviation and miss more details of the input image as shown in Fig.~\ref{fig:ofa}(a)-(b), leading to decreasing evaluation metrics.
\textbf{(2) Text Segmentation:}
As shown in Fig.~\ref{fig:ofa}(c)-(d), OFA struggles to generate the complex patterns of texts at the stroke level, primarily limited by the reconstruction capacity of the VQGAN decoder.
Moreover, as the network goes deeper, the less fine-grained alignment with the input image can be guaranteed.
It can be seen that OFA$_{huge}$ just draws text-like but meaningless patterns. 
\textbf{(3) Tampered Text Detection:}
With the indirect supervision of token sequences with condensed information, OFA can hardly distinguish the inconspicuous difference between real and tampered texts, resulting in significantly poor performance on tampered text detection (Tab.~\ref{tab:ofa}).
Moreover, the OFA is likely to produce visually plausible patches of red and green colors, ignoring the real text distribution of the input image, as shown in Fig.~\ref{fig:ofa}(e)-(f).
Finally, it is worth noting that the proposed UPOCR substantially outperforms existing generalist approaches based on either sequence-to-sequence or image-to-image paradigms, demonstrating its remarkable effectiveness and promising future in building unified pixel-level OCR interfaces.

\begin{table*}
    \centering
    \caption{Comparison with \textbf{generalist} models on pixel-level OCR tasks.
            (Params: Parameters)}
    \label{tab:ofa}
    \resizebox{\columnwidth}{!}{
        \begin{tabular}{rcccccccccccc}
            \toprule
            \multirow{2}*{Method} & \multicolumn{7}{c}{Text Removal} & \multicolumn{2}{c}{Text Segmentation} & \multicolumn{2}{c}{Tampered Text Det.} & \multirow{2}*{Params$\downarrow$}\\
            \cmidrule(r){2-8} \cmidrule(r){9-10} \cmidrule(r){11-12}
            & PSNR$\uparrow$ & MSSIM$\uparrow$ & MSE$\downarrow$ & AGE$\downarrow$ & pEPs$\downarrow$ & pCEPs$\downarrow$ & FID$\downarrow$ & fgIoU$\uparrow$ & F$\uparrow$ & mIoU$\uparrow$ & mF$\uparrow$ \\
            \midrule 
            \multicolumn{13}{c}{Sequence-to-Sequence-Based Generalist Model} \\
            \midrule 
            OFA$_{tiny}$~\cite{wang2022ofa} & 20.07 & 67.34 & 1.2419 & 15.86 & 0.2262 & 0.1038 & 68.46 & 48.40 & 65.23 & 14.70 & 25.17 & \textbf{33M} \\
            OFA$_{medium}$~\cite{wang2022ofa} & 19.13 & 62.38 & 1.4990 & 18.06 & 0.2639 & 0.1354 & 88.05 & 38.51 & 55.61 & 10.19 & 17.86 & \underline{93M} \\
            OFA$_{base}$~\cite{wang2022ofa} & 19.45 & 64.73 & 1.3994 & 16.88 & 0.2430 & 0.1172 & 78.85 & 47.77 & 64.66 & 6.34 & 11.90 & 182M \\
            OFA$_{large}$~\cite{wang2022ofa} & 16.04 & 58.08 & 3.0576 & 20.93 & 0.3115 & 0.1791 & 119.82 & 30.26 & 46.46 & 6.07 & 11.41 & 472M \\
            OFA$_{huge}$~\cite{wang2022ofa} & 15.73 & 55.80 & 3.2342 & 22.27 & 0.3351 & 0.1993 & 134.73 & 27.75 & 43.44 & 3.78 & 7.25 & 930M \\
            \midrule
            \multicolumn{13}{c}{Image-to-Image-Based Generalist Model} \\
            \midrule
            Painter~\cite{wang2023images} & \underline{27.13} & \underline{91.67}& \underline{0.2942} & \underline{8.68} & \underline{0.0898} & \underline{0.0425} & \underline{21.90} & \underline{86.36} & \underline{92.68} & \underline{69.26} & \underline{81.83} & 371M \\
            UPOCR (Ours) & \textbf{37.14} & \textbf{97.62} & \textbf{0.0428} & \textbf{1.72} & \textbf{0.0064} & \textbf{0.0034} & \textbf{10.47} & \textbf{88.76} & \textbf{94.04} & \textbf{71.71} & \textbf{83.53} & 108M \\
            \bottomrule
        \end{tabular}
    }
\end{table*}

\begin{table*}[t]
    \centering 
    \caption{Ablation study on different model sizes.
            (Params: Parameters)}
    \label{tab:model_size}
    \resizebox{\columnwidth}{!}{
        \begin{tabular}{lcccccccccccc}
            \toprule
            \multirow{2}*{Model Size} & \multicolumn{7}{c}{Text Removal} & \multicolumn{2}{c}{Text Segmentation} & \multicolumn{2}{c}{Tampered Text Det.} & \multirow{2}*{Params$\downarrow$} \\
            \cmidrule(r){2-8} \cmidrule(r){9-10} \cmidrule(r){11-12}
            & PSNR$\uparrow$ & MSSIM$\uparrow$ & MSE$\downarrow$ & AGE$\downarrow$ & pEPs$\downarrow$ & pCEPs$\downarrow$ & FID$\downarrow$ & fgIoU$\uparrow$ & F$\uparrow$ & mIoU$\uparrow$ & mF$\uparrow$ \\
            \midrule
            UPOCR-Tiny & 36.87 & \underline{97.59} & \underline{0.0430} & 1.77 & \underline{0.0065} & \textbf{0.0034} & 10.60 & \underline{87.33} & \underline{93.23} & 68.84 & 81.54 & \textbf{65M} \\
            UPOCR-Small & \underline{37.14} & \textbf{97.62} & \textbf{0.0428} & \underline{1.72} & \textbf{0.0064} & \textbf{0.0034} & \textbf{10.47} & \textbf{88.76} & \textbf{94.04} & \underline{71.71} & \underline{83.53} & \underline{108M} \\
            UPOCR-Base & \textbf{37.16} & \textbf{97.62} & 0.0451 & \textbf{1.68} & 0.0066 & \underline{0.0035} & \underline{10.54} & 87.20 & 93.16 & \textbf{72.95} & \textbf{84.36} & 192M \\
            \bottomrule
        \end{tabular}
    }
\end{table*}

\noindent\textbf{Training Details of Painter and OFA.} 
Following the training setting of UPOCR as specified in Sec.~\ref{sec:train_setting}, the Painter (Sec.~\ref{sec:exp_generalist} of the main paper) and OFA are trained for 80,000 iterations with a batch size of 48.
Moreover, the text removal, text segmentation, and tampered text detection tasks each occupy 16 samples of a batch.
Other training settings are kept the same as their original implementations.

\section{Ablation Study on Model Size}
As described in Sec.~\ref{sec:exp_set} of the main paper, the encoder-decoder architecture of UPOCR inherits from ViTEraser-Small~\cite{peng2024viteraser}.
In Tab.~\ref{tab:model_size}, we further investigate the effect of model size on the generalist capabilities of UPOCR.
In addition to UPOCR (denoted as UPOCR-Small for clarity), we build UPOCR-Tiny and UPOCR-Base following the encoder-decoder architectures of ViTEraser-Tiny and ViTEraser-Base, respectively.
The UPOCR-Tiny and UPOCR-Base are trained following the same setting as UPOCR and also use the weights of corresponding ViTEraser as initialization.
It can be seen that UPOCR-Small significantly outperforms UPOCR-Tiny on all three tasks.
However, UPOCR-Base with a larger model size exhibits inferior performance than UPOCR-Small on text segmentation and most metrics of text removal, probably due to insufficient training samples.
Therefore, we opt for UPOCR-Small as the final implementation of the proposed UPOCR.

\begin{table*}[t]
    \centering 
    \caption{Ablation study on different weight initializations during training.}
    \label{tab:init}
    \resizebox{0.7\columnwidth}{!}{
    \begin{tabular}{lcccccccc}
        \toprule
        \multirow{2}*{Initialization} & \multicolumn{4}{c}{Text Removal} & \multicolumn{2}{c}{Text Segmentation} & \multicolumn{2}{c}{Tampered Text Det.} \\
        \cmidrule(r){2-5} \cmidrule(r){6-7} \cmidrule(r){8-9}
        & PSNR$\uparrow$ & MSSIM$\uparrow$ & MSE$\downarrow$ & FID$\downarrow$ & fgIoU$\uparrow$ & F$\uparrow$ & mIoU$\uparrow$ & mF$\uparrow$ \\
        \midrule
        ImageNet & 35.67 & \underline{96.86} & 0.3556 & 12.76 & 86.53 & 92.78 & 47.96 & 64.80 \\
        SegMIM & \underline{37.04} & \textbf{97.62} & \underline{0.0433} & \underline{10.64} & \underline{88.53} & \underline{93.92} & \textbf{73.62} & \textbf{84.80} \\
        ViTEraser & \textbf{37.14} & \textbf{97.62} & \textbf{0.0428} & \textbf{10.47} & \textbf{88.76} & \textbf{94.04} & \underline{71.71} & \underline{83.53} \\
        \bottomrule
    \end{tabular}
    }
\end{table*}

\section{Ablation Study on Weight Initialization}
Retrospectively, existing generalist models~\cite{li2023blip,alayrac2022flamingo,liu2023visual,lv2023kosmos,zhu2024minigpt} mostly rely on pretrained weights from strong vision or language models.
As described in Sec.~\ref{sec:exp_set} of the main paper, UPOCR uses the pretrained \textit{ViTEraser} (with SegMIM pre-training)~\cite{peng2024viteraser} for initialization during training.
In Tab.~\ref{tab:init}, other two ways of weight initialization are investigated, including \textit{ImageNet} and \textit{SegMIM}.
Specifically, for \textit{ImageNet} manner, a Swin Transformer v2~\cite{liu2022swin} pretrained on the ImageNet-1k~\cite{deng2009imagenet} classification task is adopted to initialize the encoder, while for \textit{SegMIM} manner, pretrained weights using SegMIM~\cite{peng2024viteraser} are employed to initialize the encoder-decoder.
It can be seen that the initialization is critical to the generalist model performance. 
The \textit{SegMIM} and \textit{ViTEraser} pretrained weights may effectively learn the essential capacity to grasp text shapes and textures as well as the distinctive features between texts and backgrounds.
Therefore, the UPOCR can be rapidly adapted to various pixel-level OCR tasks with superior performance.
Although the \textit{SegMIM} is also an effective pre-training approach for UPOCR, we opt for the \textit{ViTEraser} pretrained weights due to the better performance on two of the three tasks.

\begin{table*}[t]
    \centering 
    \caption{Ablation study on feature loss.}
    \label{tab:feature_loss}
    \resizebox{0.7\columnwidth}{!}{
    \begin{tabular}{ccccccccc}
        \toprule
        \multirow{2}*{Feature Loss} & \multicolumn{4}{c}{Text Removal} & \multicolumn{2}{c}{Text Segmentation} & \multicolumn{2}{c}{Tampered Text Det.} \\
        \cmidrule(r){2-5} \cmidrule(r){6-7} \cmidrule(r){8-9}
        & PSNR$\uparrow$ & MSSIM$\uparrow$ & MSE$\downarrow$ & FID$\downarrow$ & fgIoU$\uparrow$ & F$\uparrow$ & mIoU$\uparrow$ & mF$\uparrow$ \\
        \midrule
        × & 36.99 & \textbf{97.83} & 0.0433 & 12.86 & 88.69 & 94.01 & 71.42 & 83.10 \\
        \checkmark & \textbf{37.14} & 97.62 & \textbf{0.0428} & \textbf{10.47} & \textbf{88.76} & \textbf{94.04} & \textbf{71.71} & \textbf{83.53} \\
        \bottomrule
    \end{tabular}
    }
\end{table*}

\section{Ablation Study on Feature Loss}

As specified in Sec.~\ref{sec:unified_training_strategy}, the training strategy of UPOCR is unified as minimizing the distance between predicted and GT images at pixel and feature spaces despite the heterogeneity within various pixel-level OCR tasks.
In Tab.~\ref{tab:feature_loss}, we further investigate the effect of feature loss on the versatile capacity of UPOCR.
It can be seen that removing feature loss leads to a decline in the performance of all three tasks.
Especially for the FID metric of text removal which measures the similarity between predicted and GT images, the performance drops a lot from 10.47 to 12.86.
For the text removal task, the feature loss diminishes the disparity between predicted and GT images from a human-like perceptual perspective, thereby enhancing the verisimilitude of text-erased outputs.
Moreover, the high-level semantic constraints acquired by feature alignment can facilitate the perception of text patterns and fine-grained textures, which empowers the model to precisely distinguish texts from backgrounds and capture inconspicuous differences between real and tampered texts.

\begin{table*}[t]
    \centering 
    \caption{Ablation study on task prompt.}
    \label{tab:task_prompt}
    \resizebox{0.7\columnwidth}{!}{
    \begin{tabular}{ccccccccc}
        \toprule
        \multirow{2}*{Task Prompt} & \multicolumn{4}{c}{Text Removal} & \multicolumn{2}{c}{Text Segmentation} & \multicolumn{2}{c}{Tampered Text Det.} \\
        \cmidrule(r){2-5} \cmidrule(r){6-7} \cmidrule(r){8-9}
        & PSNR$\uparrow$ & MSSIM$\uparrow$ & MSE$\downarrow$ & FID$\downarrow$ & fgIoU$\uparrow$ & F$\uparrow$ & mIoU$\uparrow$ & mF$\uparrow$ \\
        \midrule
        × & 36.33 & 96.34 & 0.641 & 13.22 & 83.59 & 91.06 & 43.39 & 60.37 \\
        \checkmark & \textbf{37.14} & \textbf{97.62} & \textbf{0.0428} & \textbf{10.47} & \textbf{88.76} & \textbf{94.04} & \textbf{71.71} & \textbf{83.53} \\
        \bottomrule
    \end{tabular}
    }
\end{table*}

\begin{figure*}[t]
    \centering
    \includegraphics[width=0.8\columnwidth]{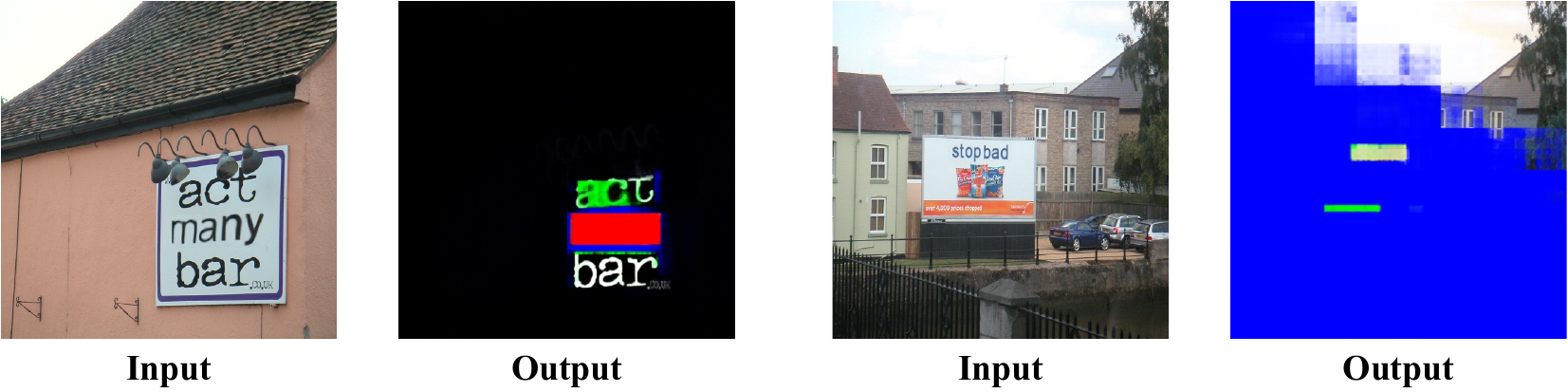}
    \caption{Visualization results of UPOCR on Tampered-IC13 without task prompts. 
    The output images may comprise a mixture of predicted pixels for divergent tasks, indicating the model is unaware of the target task.}
    \label{fig:without_task_prompt}
\end{figure*}

\section{Ablation Study on Task Prompt}

The learnable task prompts are indispensable to the generalist capability of UPOCR.
Specifically, they are injected into the ViT-based encoder-decoder to push the general representations extracted by the encoder towards task-specific regions, enabling the decoder to generate output images for the target task.
In Tab.~\ref{tab:task_prompt}, we present the performance of UPOCR with and without task prompts.
For the experiment without task prompts, we remove task prompts from UPOCR and re-train the model with the same experimental settings as described in Sec.~\ref{sec:train_setting}.
The experimental results demonstrated that the UPOCR with prompts substantially outperforms the counterpart without prompts. 
Moreover, it is interesting that the performance of the model without prompts is not extremely low. 
However, this phenomenon does not indicate the model can actually be aware of the target task. 
Because the three datasets adopted in our experiments are not \textit{i.i.d.}, the model can implicitly recognize which dataset the samples belong to and perform the corresponding task, which is known as the dataset bias problem~\cite{liu2024decade}. 
Nevertheless, the samples of Tampered-IC13 are sourced from general text spotting datasets and severely overlap with the other two datasets in terms of styles and scenarios; therefore, the model cannot accurately classify these samples into correct datasets and perform correct tasks, resulting in much lower performance. 
The visualizations in Fig.~\ref{fig:without_task_prompt} also qualitatively showcase that the model exhibits severe confusion about the target task due to the absence of task prompts.
Moreover, we cannot perform multiple tasks on one image as shown in Fig.~\ref{fig:task_prompt} if without task prompts, because the output is deterministic for the same input in this case. 
In contrast, the proposed task prompts can prevent the model from implicit dataset classification and effectively guide the model to perform diverse tasks.

\section{Speed}
The inference speed of UPOCR is 17fps using an NVIDIA RTX 3090 GPU and 2.5fps using an Intel Xeon Platinum 8375C CPU. 
Both the speeds on GPU and CPU are tested directly with PyTorch implementation and a batch size of 1.

\section{Comparison with ViTEraser}
Although the encoder-decoder architecture of UPOCR is implemented following ViTEraser~\cite{peng2024viteraser}, the proposed UPOCR is a brand-new approach that significantly differs from ViTEraser in the following aspects.
\begin{itemize}
    \item The UPOCR aims to build a generalist model for the unified pixel-level OCR interface. 
    To achieve this, UPOCR unifies the paradigm, architecture, and training strategy of diverse pixel-level OCR tasks.
    Comprehensive experiments demonstrate the state-of-the-art performance of UPOCR in simultaneously handling text removal, text segmentation, and tampered text detection tasks with a single unified model.
    In contrast, ViTEraser specializes in the text removal task without generalist capacities.
    Moreover, it relies on dedicated modules including a discriminator and an auxiliary mask branch and is trained following a complicated GAN-based strategy.
    Although ViTEraser can be extended to the tampered text detection task, its paradigm and training strategy are adapted from image generation to segmentation and its parameters are re-trained using the tampered text detection dataset.
    \item To obtain the multi-task processing ability, UPOCR introduces learnable task prompts into ViT-based encoder-decoder architecture in a simple-yet-effective fashion.
    The task prompt pushes the general representation extracted by the encoder towards task-specific regions, empowering the decoder to generate output images for individual tasks.
    \textit{Although we borrow the encoder-decoder architecture of ViTEraser to instantiate UPOCR, it is irrelevant to the novel part of our study.}
    Additionally, the U-shaped ViT-based architecture of ViTEraser itself is a general framework which has been broadly investigated arcoss multiple domains~\cite{cao2022swin,wang2022uformer}.  
    \item In this paper, extensive experiments and analyses are conducted to provide deep insights into the construction of unified pixel-level OCR interfaces, which we believe could spark more future research on generalist OCR models. 
\end{itemize}

\section{Limitation}
Despite the effectiveness of the proposed UPOCR as a unified pixel-level OCR interface, the limitation lies in the generalization ability to unseen tasks.
Although the learnable task prompts empower the UPOCR to simultaneously handle multiple tasks, they are fixed once the model training is finished, which means re-training is necessary if the pre-set range of tasks is expanded.
Since task prompts function as feature-level offsets that push general representations to task-specific spaces, the automatic learning of prompts from example input-output pairs of new tasks is critical to a flexible generalist pixel-level OCR model, which is worth studying in future work.

\section{More Visualization}
\begin{figure*}
    \centering 
    \includegraphics[width=\columnwidth]{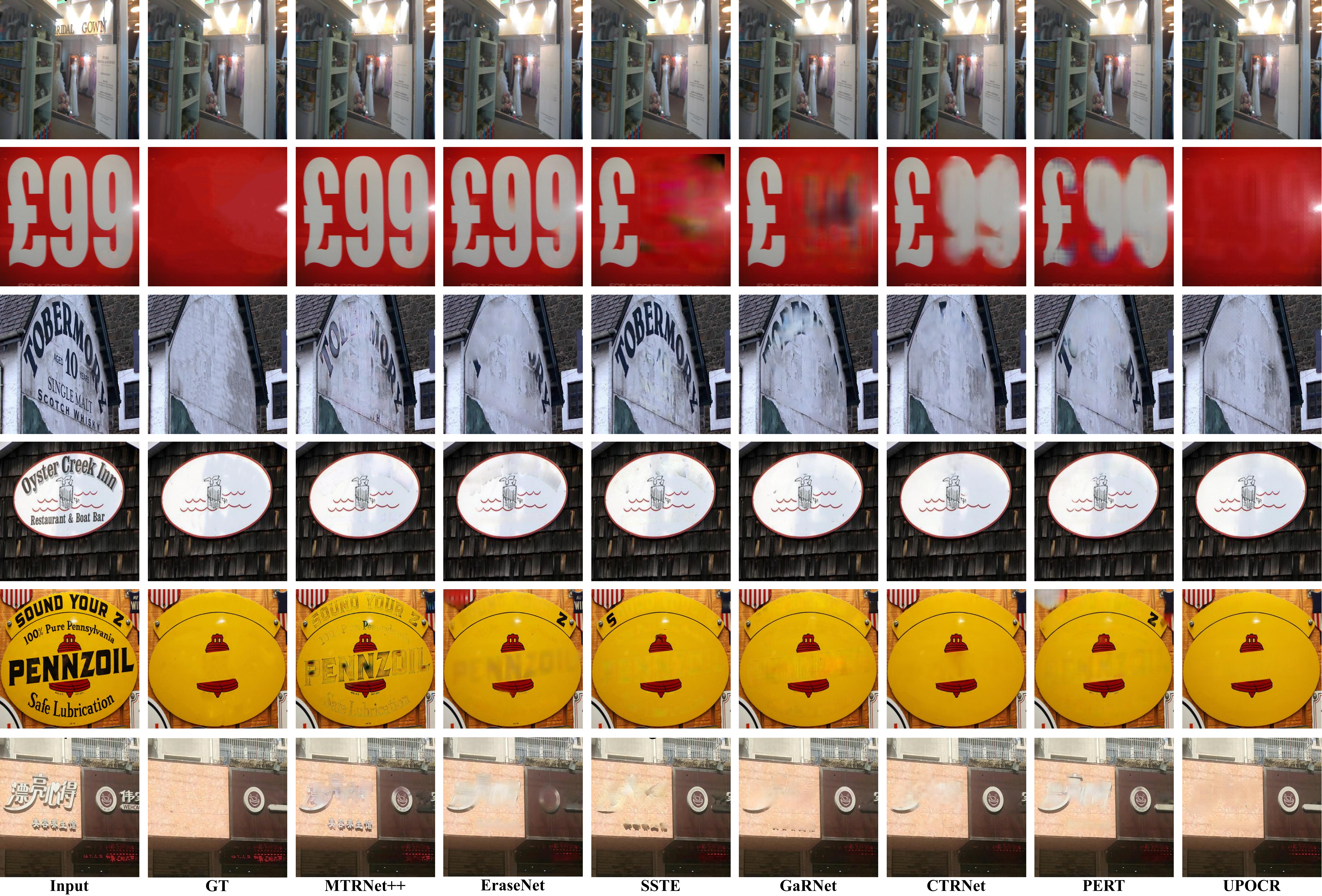}
    \caption{More visualizations on text removal,
    where the inference results are obtained by MTRNet++~\cite{tursun2020mtrnet++}, EraseNet~\cite{liu2020erasenet}, SSTE~\cite{tang2021stroke}, GaRNet~\cite{lee2022surprisingly}, CTRNet~\cite{liu2022don}, PERT~\cite{wang2023pert}, and UPOCR (ours).
    Zoom in for a better view.}
    \label{fig:text_removal}
\end{figure*}

\begin{figure*}
    \centering 
    \includegraphics[width=\columnwidth]{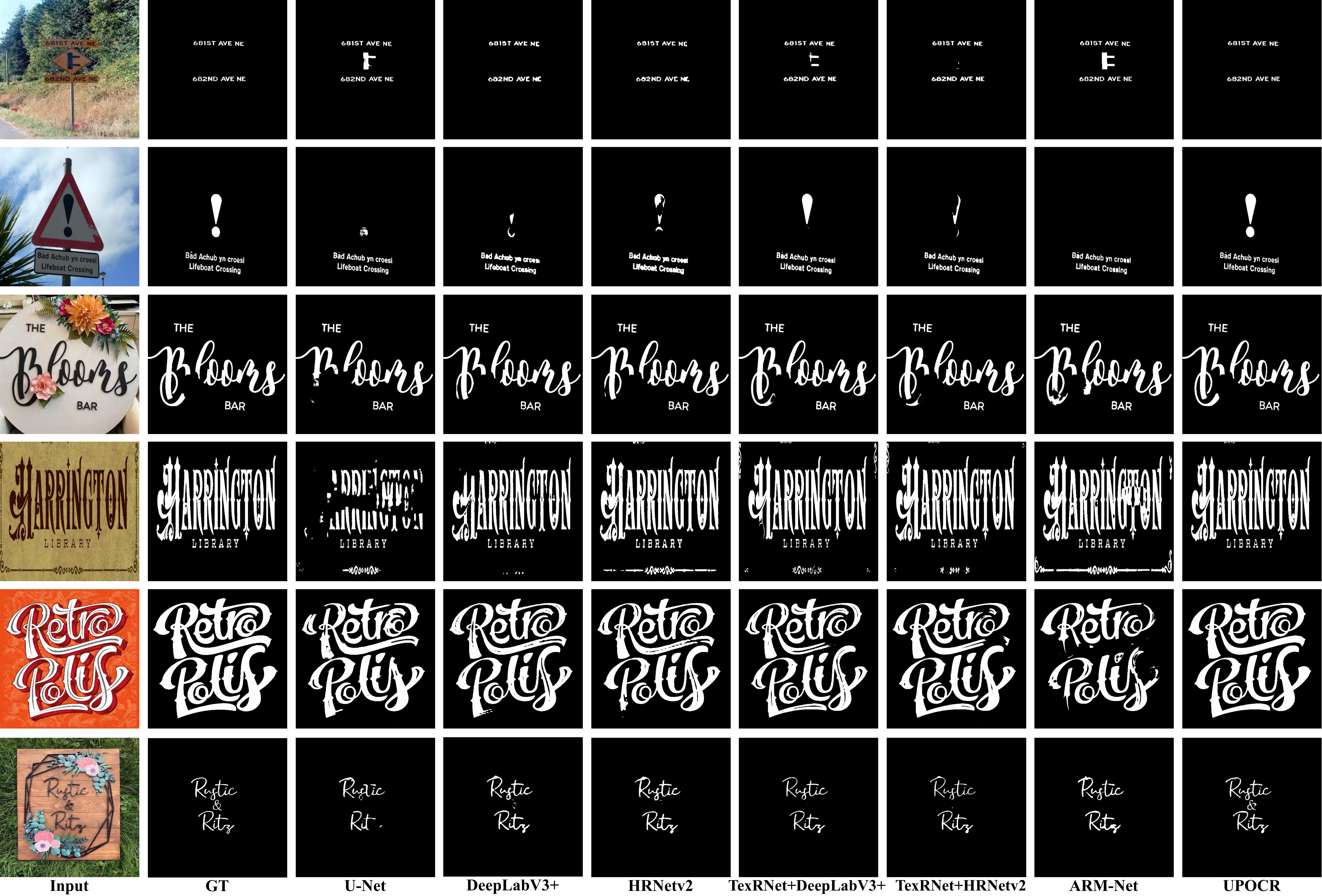}
    \caption{More visualizations on text segmentation,
    where the inference results are obtained by U-Net~\cite{ronneberger2015u}, DeepLabV3+~\cite{chen2018encoder}, HRNetv2~\cite{wang2020deep}, TexRNet+DeepLabV3+~\cite{xu2021rethinking}, TexRNet+HRNetv2~\cite{xu2021rethinking}, ARM-Net~\cite{ren2022looking}, and UPOCR (ours).
    Zoom in for a better view.}
    \label{fig:text_seg}
\end{figure*}

\begin{figure*}
    \centering 
    \includegraphics[width=\columnwidth]{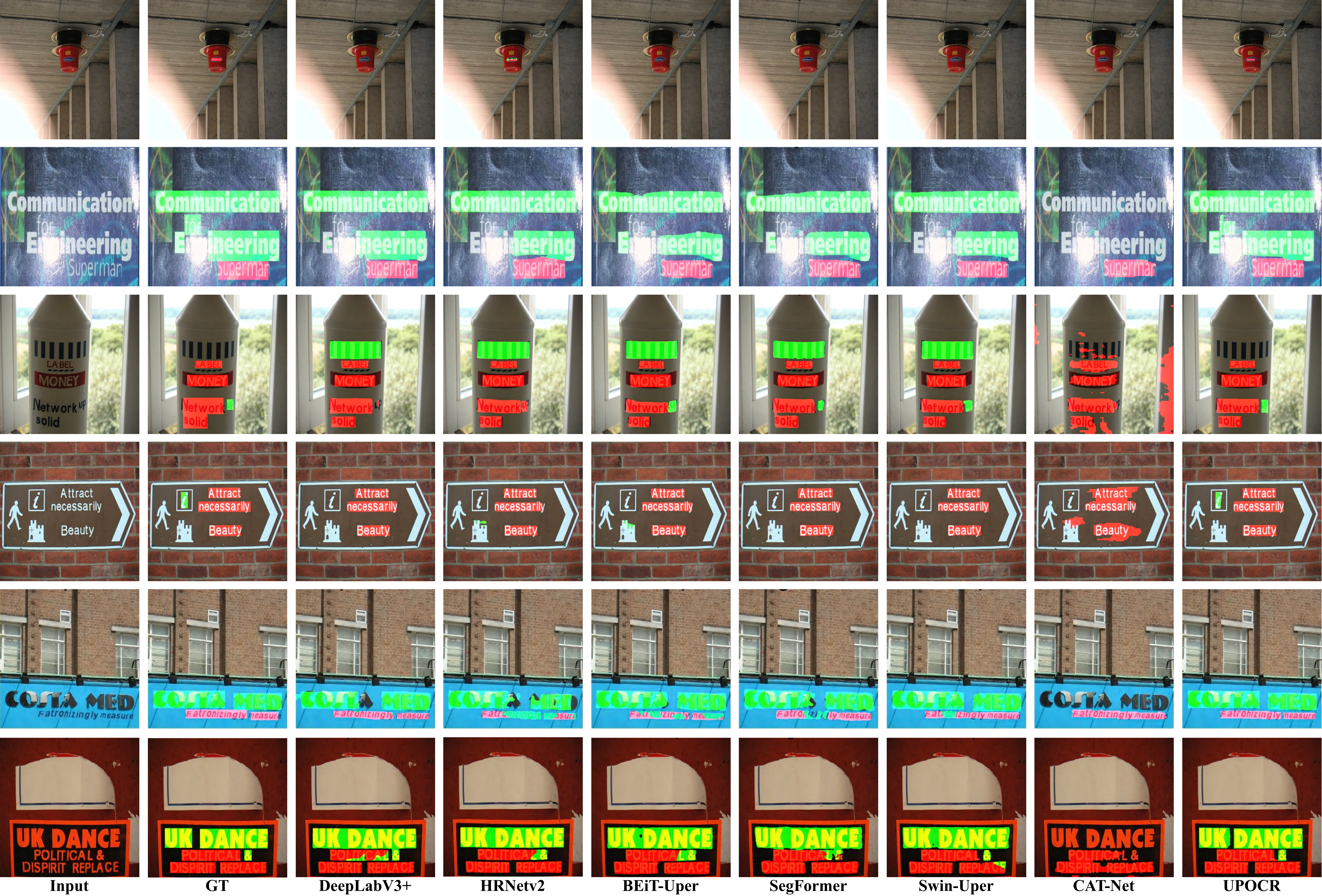}
    \caption{More visualizations on tampered text detection, where the inference results are obtained by DeepLabV3+~\cite{chen2018encoder}, HRNetv2~\cite{wang2020deep}, BEiT-Uper~\cite{bao2022beit}, SegFormer~\cite{xie2021segformer}, Swin-Uper~\cite{liu2021swin}, CAT-Net~\cite{kwon2022learning}, and UPOCR (ours).
    Zoom in for a better view.}
    \label{fig:ttd}
\end{figure*}

In this section, we provide more qualitative results on text removal, text segmentation, and tampered text detection tasks in Figs.~\ref{fig:text_removal}, \ref{fig:text_seg}, and \ref{fig:ttd}, respectively.
It can be seen that the proposed UPOCR simultaneously excels in multiple tasks using a unified single model without task- or benchmark-specific finetuning.
Especially in the first row in Fig.~\ref{fig:text_removal}, our method demonstrates superiority in tackling tiny and densely distributed text.

\begin{figure*}
    \centering 
    \includegraphics[width=0.7\columnwidth]{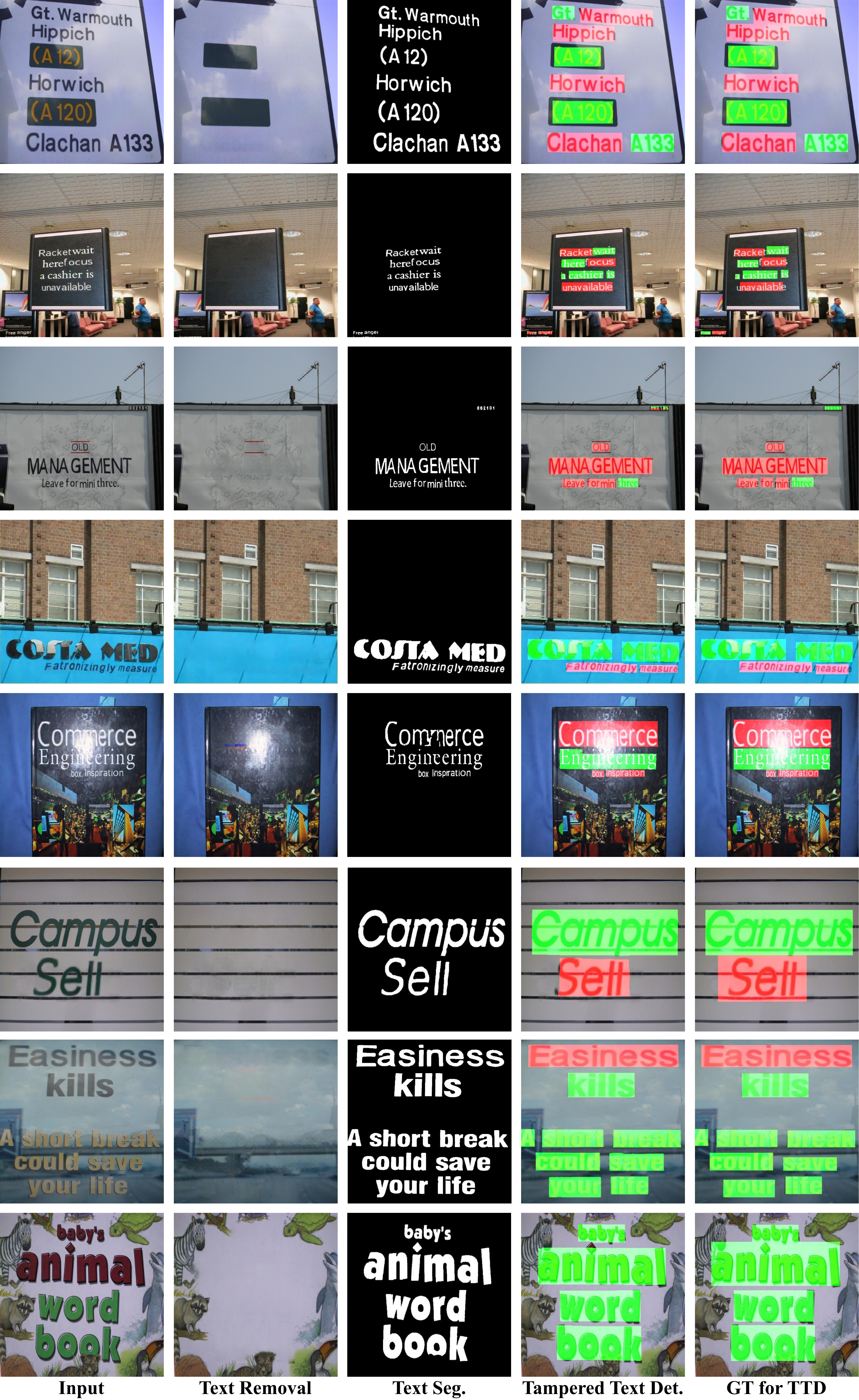}
    \caption{More visualizations of three-task outputs on Tampered-IC13 dataset. 
    All tasks are performed with a single UPOCR model using corresponding task prompts.
    The GT of tampered text detection (TTD) provided by Tampered-IC13 is also presented for comparison.
    Zoom in for a better view.
    (Seg.: Segmentation, Det.: Detection)}
    \label{fig:three_task}
\end{figure*}

Moreover, in addition to Fig.~\ref{fig:task_prompt} of the main paper, we supplement more visualizations of three-task outputs using the test set of Tampered-IC13~\cite{wang2022detecting} in Fig.~\ref{fig:three_task}.
Specifically, all tasks are conducted with a single UPOCR model using corresponding task prompts.
As demonstrated by the visualizations, UPOCR can effectively perform diverse tasks on the same input image, indicating the proposed learnable task prompts adequately guide the generation process of the decoder.

\end{document}